\documentclass[journal]{IEEEtran}
\usepackage{amsmath,graphicx}
\usepackage{times}
\usepackage{epsfig}
\usepackage{amssymb}
\usepackage{amsmath}
\usepackage{subcaption}
\usepackage{enumitem}
\usepackage{booktabs}
\usepackage{multirow}
\usepackage[noadjust]{cite}

\usepackage[pagebackref=true,breaklinks=true,colorlinks,bookmarks=false]{hyperref}

\DeclareMathOperator{\sign}{sgn}

\DeclareMathOperator*{\argmin}{arg\,min}

\ifCLASSINFOpdf
\else
\fi
\usepackage{amsmath}
\allowdisplaybreaks
\begin{document}
%
\title{Level Set Binocular Stereo with Occlusions}
%
%
%

\author{Jialiang~Wang,~\IEEEmembership{Student Member,~IEEE,}
        and~Todd~Zickler,~\IEEEmembership{Member,~IEEE}
\thanks{JW and TZ are with John A. Paulson School of Engineering and Applied Sciences, Harvard University, Allston,
MA, 02134 USA}
\thanks{Earlier versions of this manuscript appeared at the International Conference on Image Processing~\cite{wang2020level} and in Chapters 1 and 5 of JW's PhD dissertation~\cite{wangthesis}. This version contains more detailed analysis, additional experiments with different parameters, and additional derivations and illustrations.}%
\thanks{e-mails: jialiang.wang@alumni.harvard.edu; zickler@seas.harvard.edu.}
}

%
%

\markboth{}%
{Wang \MakeLowercase{\textit{et al.}}: Level Set Binocular Stereo with Occlusions}
%



\maketitle

\begin{abstract}
Localizing stereo boundaries and predicting nearby disparities are difficult because stereo boundaries induce occluded regions where matching cues are absent. Most modern computer vision algorithms treat occlusions secondarily (e.g., via left-right consistency checks after matching) or rely on high-level cues to improve nearby disparities (e.g., via deep networks and large training sets). They ignore the geometry of stereo occlusions, which dictates that the spatial extent of occlusion must equal the amplitude of the disparity jump that causes it. 
This paper introduces an energy and level-set optimizer that improves boundaries by encoding occlusion geometry. Our model applies to two-layer, figure-ground scenes, and it can be implemented cooperatively using messages that pass predominantly between parents and children in an undecimated hierarchy of multi-scale image patches. In a small collection of figure-ground scenes curated from Middlebury and Falling Things stereo datasets, our model provides more accurate boundaries than previous occlusion-handling stereo techniques. This suggests new directions for creating cooperative stereo systems that incorporate occlusion cues in a human-like manner.

\end{abstract}
%


\section{Introduction}
\label{sec:intro}


Deep convolutional networks can provide fast and accurate estimates of binocular stereo disparity by internalizing and exploiting local and non-local patterns of scene shape and appearance in a dataset~\cite{poggi2020synergies}, but their reliance on spatial sub-sampling (i.e., stride and pooling) limits their accuracy near object boundaries~\cite{wang2019local,scharstein2014high}. One way to address this is to develop iterative bottom-up systems for disparity estimation that can eventually be combined with fast, feed-forward estimates from a CNN, and that complement their coarse top-down disparity information by analyzing local disparity signals at high spatial resolution and explicitly modeling the smooth curvilinear properties of boundaries.

One appealing class of bottom-up techniques for estimating boundaries in two-dimensional signals are active contours implemented as level sets. These have the benefits of not presupposing a boundary topology, and of being able to simultaneously exploit the smooth curvilinear behavior of boundaries while also exploiting the smoothness of signals in between them~\cite{caselles1997geodesic,chan2001active}. 

A substantial challenge in creating level set boundary techniques for binocular disparity signals is that each disparity boundary has an adjacent region where the matching signal is ``missing''. A foreground object necessarily causes an adjacent spatial region of the background to be visible in only one of the left or right input images (Fig.~\ref{supfig:45_geometry}(a)), and this causes the left-right disparity matching signal to be invalid in that occluded region (Fig.~\ref{supfig:45_geometry}(b)). 

In this paper, we take a step toward a level-set framework for binocular stereo boundaries by examining the special case of figure-ground scenes comprising two depth layers: a foreground layer and a background layer. Our key contributions are an energy and multiscale optimization strategy that accurately hard-code the geometry of occlusions, namely that in a rectified stereo pair, the spatial extent of an occluded region is equal to the magnitude of the disparity discontinuity that causes it~\cite{belhumeur1996bayesian}. This constraint has been properly enforced in some one-dimensional algorithms that operate on isolated stereo scanlines~\cite{belhumeur1996bayesian,bobick1999large,wang2017toward}, but it has been ignored or crudely approximated in previous two-dimensional stereo algorithms, including those based on level sets~\cite{deriche1997level}. 


Another benefit of our approach is that it can be implemented cooperatively using local calculations among distributed computational units. The units have overlapping receptive fields at multiple scales, and each unit maintains a compact local state and shares information through only a sparse set of connections in location and scale. In this sense, the model exhibits some basic tenets of biological plausibility. 

We evaluate our model by measuring the accuracy of estimated disparity and occlusions at and near foreground/background boundaries in a  small collection of synthetic and captured images curated from the Middlebury~\cite{scharstein2007learning,hirschmuller2007evaluation} and Falling Things~\cite{tremblay2018falling} datasets. With approximate initialization, our model converges to estimates of foreground disparity boundaries and background occlusions that are more accurate than those of existing techniques.   



Our exposition begins by revisiting the geometry of stereo occlusions, as well as the concepts of disparity, stereo cost volumes derived from matching, and the local information that exists within these cost volumes for the locations of foreground disparity boundaries (Sec.~\ref{sec:45degree}). We follow these preliminaries with a summary of related work (Sec.~\ref{sec:related}), and then we introduce and evaluate our objective and optimization strategy, which are based on a level set formulation that discounts matching penalties in regions of occlusion. 


\section{Preliminaries}
\label{sec:45degree}

\begin{figure*}[h!]
    \centering
    \includegraphics[width=0.75\textwidth]{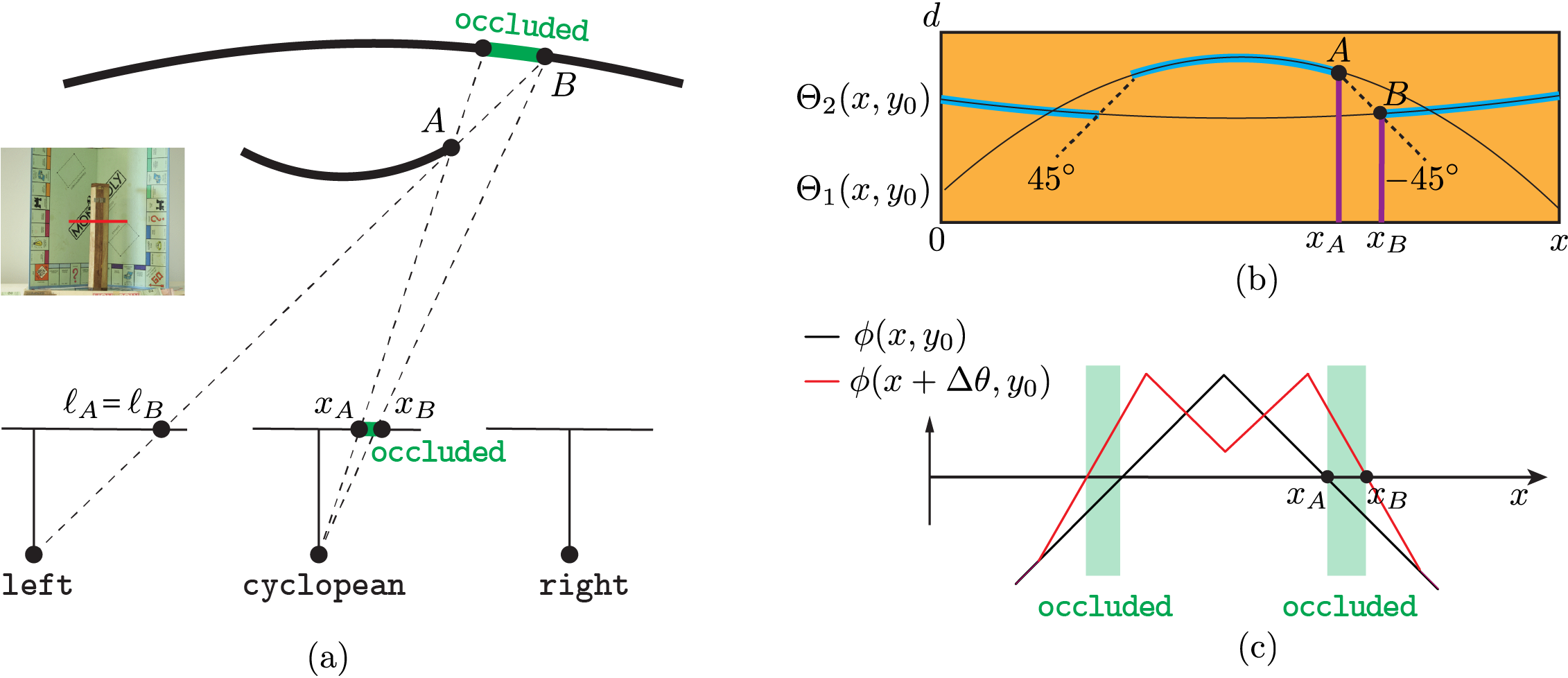}
    \caption{Occlusion geometry. (a) Cross-section of scene and cameras in an epipolar plane (with $y=y_o)$. Point $A$ is a foreground occluding boundary, and point $B$ is the boundary of the occluded background. (b) Corresponding slice of cost volume, with matching cost abstracted as high (orange) or low (blue). Global models $\Theta_1(x, y_o), \Theta_2(x, y_o)$ are superimposed. Point $B$ in disparity space must lie on a ray from point $A$ with slope $-45^\circ$. (c) We encode this geometry using intermediate function $\Delta\theta(x,y_o;\mathbf{\Theta}_1, \mathbf{\Theta}_2)$ from Eq.~\ref{supeqn:delta_theta}, so occluded regions (green) statisfy $\phi(x,y) < 0$ and $\phi(x+\Delta\theta(x,y;\mathbf{\Theta}_1, \mathbf{\Theta}_2),y)>0$.}
    \label{supfig:45_geometry}
\end{figure*}

As is typical, we assume left and right input images are rectified to have aligned horizontal epipolar scanlines, which corresponds to having two effective cameras with equal focal lengths and aligned, parallel image planes. See Fig.~\ref{supfig:45_geometry}(a).
We follow~\cite{belhumeur1996bayesian} by defining a virtual cyclopean camera that is centered between the rectified left and right cameras and that shares the same horizontal epipolar scanlines. We represent disparity, which is inversely proportional to depth, as a function on the cyclopean visual field $(x,y)\in\Omega\subset\mathbb{R}^2$, where $x$ indexes locations within a scanline and $y$ indexes the set of scanlines. Our task is to associate a disparity value $d \in [0,d_{\max}]$ with each $(x,y)$, where we  define disparity as the horizontal difference (say, $\ell_A-x_A$ in Fig.~\ref{supfig:45_geometry}(a)) between a cyclopean image point (say, $x_A$) and some left image point (say, $\ell_A$).


An important signal for estimating the disparity $d$ at cyclopean point $(x,y)$ is the similarity between the values of the left image near to $(x+d,y)$ and the values of the right image near to $(x-d,y)$. We follow common practice by assuming the existence of a precomputed data structure called a stereo cost volume $C(x,y,d)\in[0,1]$ which records these as local dissimilarity scores or \emph{matching costs}. Fig.~\ref{supfig:45_geometry}(b) depicts a cartoon $(x,d)$-slice of a stereo cost volume that might exist when surface textures are highly distinctive, with matching costs abstracted as being either high or low (resp.~orange or blue). Away from surface boundaries, it is reasonable to expect that true disparity values have relatively low matching costs.

For the purposes of this paper, we restrict our estimated disparity functions to be piecewise smooth. Specifically, we define global basis functions $\mathbf{U}(x,y) = \{U_i(x,y)\}_{i=1 \cdots m}$, and within the $j$th spatial region we express the estimated disparity function as a linear combination $\Theta_j(x,y)= \mathbf{\Theta}_j \mathbf{U}(x,y) = \sum\limits_{i=1}^m \mathbf{\Theta}_j(i) U_i(x,y)$ with shape coefficients $\mathbf{\Theta}_j\in\mathbb{R}^m$. We refer to $\mathbf{\Theta}_j$ as \emph{global shapes} and use the convention $\mathbf{\Theta}_1$ and $\mathbf{\Theta}_2$ for foreground and background, respectively. Our experiments use a second-order polynomial basis, $\mathbf{U}(x,y) = \{x^2, xy, y^2, x, y, 1 \}$, and an example is depicted in Fig.~\ref{supfig:45_geometry}(b).


These conventions and definitions make the geometry of occlusions very simple~\cite{belhumeur1996bayesian}. Suppose $A$ is a foreground surface boundary as depicted in Fig.~\ref{supfig:45_geometry}(a), and suppose $B$ is the boundary of a background surface that is not visible in the left camera. The \emph{occluded region} in the visual field is the interval $(x_A,x_B)$ between their cyclopean projections.
Since the foreground and background disparity functions are smooth, we can denote them along scanline $y_o$ by $\Theta_1(x, y_o)$ and $\Theta_2(x, y_o)$, respectively, and write
\begin{equation}
\begin{split}
    \Theta_1(x_A, y_o)  &= \ell_A - x_A  \\
    \Theta_2(x_B, y_o)  &= \ell_B - x_B.    
\end{split}
\end{equation}
Subtracting these equations yields 
\begin{equation}
\begin{split}
    \Theta_1(x_A, y_o) - \Theta_2(x_B, y_o) &= x_B - x_A,
\end{split}
\end{equation}
which shows that \emph{the size of the occluded region is equal to the disparity change that occurs at the foreground/background boundary}. As in Fig.~\ref{supfig:45_geometry}(b), this can be visualized as an ``occluding ray'' with slope $-45^\circ$ in the cost volume: As the shape or position of the background surface changes, the location of $B$ in the cost volume travels along this ray. Note that for an occlusion event on the other side of a foreground object, the analogous ray has a $45^\circ$-slope of opposite sign.

There are generally two types of signals to help identify foreground occluding points within the stereo cost volume (such as point $A$ in Fig.~\ref{supfig:45_geometry}(b)). When foreground and background surfaces have different colors or textures, the projections of the occluding point into each of the left and right images will coincide with a detectable color or texture boundary in that image. We define a \emph{monocular boundary cost} $B_{\text{m}}(x,y,d)$ and assume it has been pre-computed by: (i) executing a monocular edge or boundary detector in each of the left and right images to produce left and right boundary costs $E_l(x',y'), E_r(x',y')$; and (ii) combining them using
\begin{equation} \label{eq:monocular-boundary}
    B_{\text{m}}(x,y,d) = E_l(x+d,y) + E_r(x-d,y).
\end{equation}
The second type of signal relates to the fact that $A$-like points within a cost volume have distinctive local signatures~\cite{wang2019local} associated with a rapid spatial transition from low to high matching costs (e.g., $|\partial C(x,y,d)/\partial x|$ is large~\cite{wang2017toward}). We define an \emph{occlusion boundary cost} $B_{\text{o}}(x,y,d)$ and assume it has been pre-computed by executing an occlusion detector in the cost volume. We assume both costs are normalized so $B_{\text{m}}, B_{\text{o}}\in[0,1]$ and presume that foreground occluding points tend to occur at locations with lower boundary costs.




\section{Related Work}
\label{sec:related}
Here we discuss the methods that influenced our model most. Broader reviews of stereo algorithms can be found elsewhere, including~\cite{szeliski2010computer} for classical methods and~\cite{poggi2020synergies} for methods based on deep-learning. For a summary of the biological evidence supporting the use of occlusion cues in human stereo vision, see Tsirlin et al.~\cite{tsirlin2014computational}.

\textbf{Level set stereo.} Our approach is very different from the level-set binocular stereo method of Deriche et al.~\cite{deriche1997level}, which estimates the shape of a smooth surface using a level set function in three dimensions $\phi(x,y,d)$ and does not allow the surface to self-occlude. It is also very different from level-set multi-view stereo methods (e.g.,~\cite{yezzi2003stereoscopic}), which use a volumetric level set function $\phi(x,y,z)$ to model surface shape and avoid the challenges of occlusion by assuming the input includes a sufficient number of views to guarantee that every surface point is visible in (and therefore can be matched using) at least two images. In contrast, we use a level set function in two dimensions $\phi(x,y)$ to model foreground boundaries, and we explicitly account for occlusion induced by these boundaries. 


\textbf{Stereo occlusions.} Accounting for the geometry of occlusions is well-established in scanline approaches to binocular stereo~\cite{belhumeur1996bayesian, bobick1999large, birchfield1999depth, wang2017toward}, which operate on one horizontal row at time and can use dynamic programming. But full two-dimensional stereo algorithms that exploit the curvilinear structure of boundaries across scanlines have only approximated occlusion geometry. Approaches include:
\begin{enumerate}
    \item restricting occlusion boundaries to be a subset of monocular texture and intensity boundaries~(e.g.~\cite{song2020edgestereo});

    \item treating occlusions as secondary to matching, by using a ``left-right consistency check''~\cite{weng1988two} that separately computes two disparity maps from each of the left and right viewpoints and then tries to determines occlusions from their inconsistencies;

    \item introducing an ``outlier label'' for pixels that cannot be well-matched and inferring these labels either without enforcing any occlusion geometry ~\cite{kolmogorov2001computing, ogale2005shape} or partially enforcing geometry by accounting for the polarity of a disparity jump but not its magnitude~\cite{yamaguchi2012continuous}; and

    \item augmenting left images with fake occlusions when training a CNN~\cite{yang2019hierarchical}.
\end{enumerate}
None of these methods enforce the complete occlusion geometry described in Sec.~\ref{sec:45degree}, and they all have limitations. In particular, the first one fails when occlusions do not co-occur with any texture or intensity boundary (such as in random dot stereograms~\cite{julesz1971foundations}), and the second one fails when matching cues are weak (e.g., such as the stimuli reviewed in~\cite{wang2017toward}).

\textbf{Layered stereo methods.} Our model is related to previous layer-based approaches to stereo and motion, including: early examples that introduce the idea of 
representing a disparity map as a collection of smooth  base layers plus per-pixel residual displacements within these layers (``surface + parallax'')~\cite{darrell1991robust, wang1993layered,kumar1994direct,baker1998layered}; Lin and Tomasi~\cite{lin2003surfaces}, who add a geometry-agnostic outlier label for pixels not assigned to any base layer; and Sun et al.~\cite{sun2010layered}, who further require outlier labels to respect polarity (but not magnitude) of disparity jumps between base layers. Our contribution is a descent-based method for inferring the base layers while respecting complete occlusion geometry. We do this by focusing on two-layer scenes, and we do it without explicitly estimating the per-pixel residual displacements within the inferred layers.


\textbf{Cooperative stereo methods.} 
There is a long history of cooperative stereo algorithms, from Marr and Poggio's early work~\cite{marr1976cooperative} to graph cuts~\cite{kolmogorov2001computing} and loopy belief propagation~\cite{sun2003stereo,felzenszwalb2006efficient,li2008differential}.
The most relevant to our model is Chakrabarti et al.~\cite{chakrabarti2015low}, who introduce an effective hierarchical structure of patches that we also use in our model. While some of these cooperative models allow for discounting occluded regions as outliers to matching~\cite{zitnick2000cooperative,kolmogorov2001computing,chakrabarti2015low}, none of them incorporate the geometric occlusion constraint discussed in Sec.~\ref{sec:45degree}. A notable exception is the work of Tsirlin et al.~\cite{tsirlin2014computational}, who propose a sequence of local computations that may account for these non-local constraints. 
We design a different architecture and set of dynamics from an optimization perspective.


\section{Energy and optimization}
\label{sec:method}

To obtain foreground/background boundaries, we evolve a continuous level set function $\phi(x,y)$ that is zero-valued at the boundary, positive-valued in the foreground, and negative valued in the background. It evolves in response to the three driving forces described in Sec.~\ref{sec:45degree}: stereo matching cost $C(x,y,d)$, occlusion boundary cost $B_{\text{o}}(x,y,d)$, and monocular boundary cost $B_{\text{m}}(x,y,d)$. We combine these into the energy
\begin{equation}
\begin{split}
    J(\mathbf{{\Theta}_1}, \mathbf{{\Theta}_2}, & x,y, \phi, \nabla\phi) = \int_\Omega H(\phi)C_{\Theta_1}dxdy  \\ 
  + &\int_\Omega(1-H(\phi_{+})) (1-H(\phi))C_{\Theta_2}dxdy  \\
    + & \mu \int_\Omega B_{\Theta_1} \delta(\phi) |\nabla \phi|dxdy,
\end{split}
\label{eqn:loss_function}
\end{equation}

\noindent where $H(\cdot), \delta(\cdot)$ are Heaviside and Dirac delta functions, and
{\medmuskip=3mu
\thinmuskip=2mu
\thickmuskip=4mu
\begin{align*}
    \phi & = \phi(x,y), \\
    \quad \phi_{+} & = \phi(x+\Delta\theta(x,y;  \mathbf{\Theta}_1,\mathbf{\Theta}_2),y), \\
    C_{\Theta_1} & = C(x,y,\Theta_1(x,y)), \\
    \quad C_{\Theta_2} & = C(x,y,\Theta_2(x,y)), \\
    B_{\Theta_1} & =\alpha_1 B_{\text{o}}(x,y,\Theta_1(x,y)) + \alpha_2 B_{\text{m}}(x,y,\Theta_1(x,y)) + \alpha_3
\end{align*}
}%
with tunable parameters $\alpha_i, \mu$. 

\begin{figure*}[t!]
    \centering
    \includegraphics[width=0.98\textwidth]{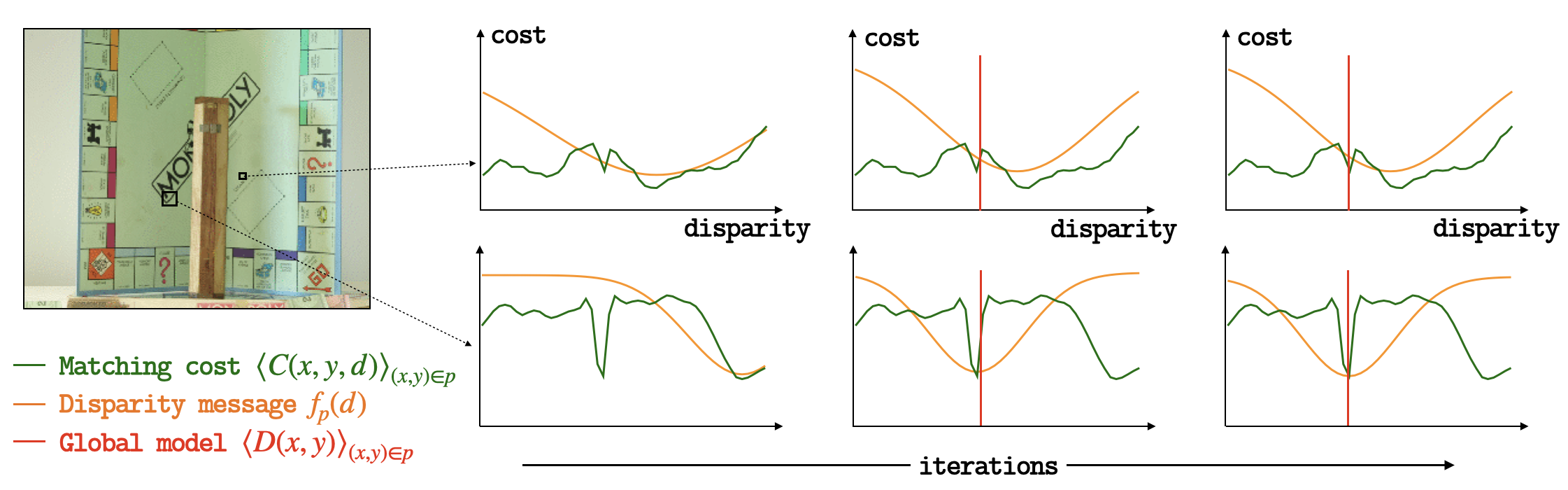}
    \caption{
    \textbf{Per-patch matching signals and dynamics}. Matching cost averaged over a single patch $p$ (green curves, via $\langle C(x,y,d)\rangle_{(x,y)\in p}$) is often unreliable, with several local minima (top) and sometimes false global minima (bottom). We equip each patch with an evolving \emph{disparity message} (orange curves, via Eq.~\ref{eq:patch-gaussian}) that balances the patch's average matching signal and the evolving global disparity map $D(x,y)$. Top: Matching signal is uniformly low due to lack of texture; disparity message persistently expresses high variance. Bottom: Message is initially erroneous but corrects over time.}
    \label{fig:patch_vote}
\end{figure*}

The first term of Eq.~\ref{eqn:loss_function} is easy to interpret  as the integrated matching cost of the foreground surface, and the third is the weighted length of the foreground boundaries with weight $B_{\Theta_1}$. The second term integrates the matching cost of the background surface, \emph{but only over the subset that is not occluded}. It uses an intermediate function $\Delta\theta(x,y; \mathbf{\Theta}_1,\mathbf{\Theta}_2)$ that tractably encodes a close approximation to the occlusion geometry of Sec.~\ref{sec:45degree}:
\begin{equation}
\label{supeqn:delta_theta}
\begin{split}
    \Delta\theta( & x,y; \mathbf{\Theta}_1,\mathbf{\Theta}_2) = \\
    & \sign\left(\frac{d\phi(x,y)}{dx}\right) \max(0, \Theta_1(x,y) - \Theta_2(x,y)),
\end{split}
\end{equation}
an example of which is in Fig.~\ref{supfig:45_geometry}(c). The $\sign()$ term distinguishes left and right sides of the foreground surface, and referring to Fig.~\ref{supfig:45_geometry}(b), the approximation comes from using the vertical distance between $A$ and $\Theta_2(x_A,y)$ as a surrogate for that between $A$ and $B$, i.e., $\Theta_1(x_A,y) - \Theta_2(x_A,y) \approx \Theta_1(x_A,y) - \Theta_2(x_B,y) = x_A - x_B$. It assumes disparity changes are small within the occluded region.

We use alternating updates to find $\mathbf{{\Theta}}_1, \mathbf{{\Theta}}_2$ and $\phi(x,y)$ that locally minimize $J$. We begin with some initialization $\phi_0(x,y)$ and at first assume no occlusion, $\Delta\theta(x,y; \mathbf{\Theta}_1,\mathbf{\Theta}_2)=0$. We alternate between: (i) updating global models $\mathbf{{\Theta}}_1, \mathbf{{\Theta}}_2$ using the current $\phi(x,y)$ and $\Delta\theta(x,y; \mathbf{\Theta}_1,\mathbf{\Theta}_2)$;
(ii) solving for the optimal shape parameters $\mathbf{{\Theta}_1}$ and $\mathbf{{\Theta}_2}$ in each region using weighted linear least squares; (iii) updating $\Delta\theta(x,y; \mathbf{\Theta}_1,\mathbf{\Theta}_2)$ using Eq.~\ref{supeqn:delta_theta}; and (iv) updating $\phi(x,y)$ by following the common  practice (e.g.,~\cite{caselles1997geodesic, chan2001active}) of replacing $\delta(\cdot)$ and $H(\cdot)$ with differentiable approximations $\delta_\epsilon(\cdot)$ and $H_\epsilon(\cdot)$ and iteratively minimizing the Euler-Lagrange equation by gradient descent. Parameterizing the descent by $t \ge 0$ and assuming that foregound surfaces are sufficiently wide (see Appendix A) one derives
\begin{equation}
\label{eqn:dphi_dt}
\begin{split}
        \frac{d\phi}{dt} = \delta_\epsilon(\phi) \biggl[ &  -  C(x,y, {\Theta}_1(x,y)) \\ 
        & + C(x-\Delta\theta(x,y;\mathbf{\Theta}_1,\mathbf{\Theta}_2),y,{\Theta}_2(x,y)) \\
        & + \mu \Big( B(x,y,{\Theta}_1(x,y)) \kappa(x,y) +  \\
        & \qquad \mathbf{N}(x,y) \cdot \nabla B(x,y,{\Theta}_1(x,y)) \Big) \biggl], 
\end{split}
\end{equation}

\noindent with  $\frac{\delta_\epsilon(\phi)}{|\nabla \phi|}\frac{d\phi}{d\vec{n}} = 0$ on the boundary $\partial \Omega$ of the visual field.
Here, $\kappa(x,y)=\text{div} (\frac{\nabla \phi(x,y)}{|\nabla \phi(x,y)|})$ and $\mathbf{N}(x,y) = \frac{\nabla \phi(x,y)}{|\nabla \phi(x,y)|}$
are the curvature and normal of the foreground contour, and $\vec{n}$ is the exterior normal to $\partial \Omega$. At any time during the evolution, a global piecewise-smooth disparity map is available via
\begin{equation}
\label{eqn:D}
    D(x,y) = H(\phi(x,y))\Theta_1(x,y) + (1-H(\phi(x,y)))\Theta_2(x,y).
\end{equation}


\section{Multiscale Alternating Descent}
\label{sec:alternating_descent}

\begin{figure*}[t!]
    \centering
    \includegraphics[width=0.98\textwidth]{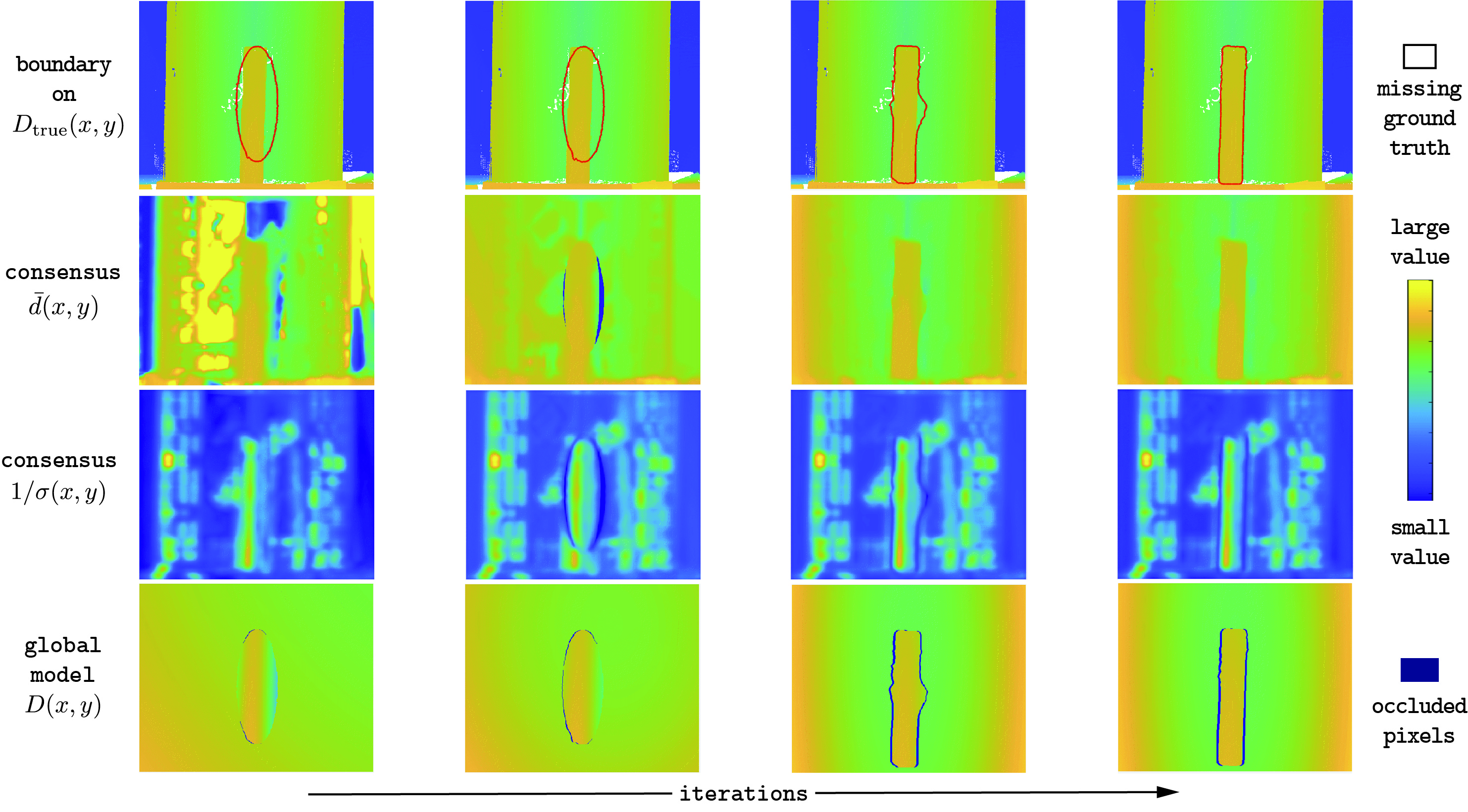}
    \caption{
    \textbf{Consensus dynamics}. From top to bottom: Evolution of boundary $\phi(x,y)=0$ superimposed on true disparity map $D_\text{true}(x,y)$; consensus mean and inverse standard deviation; and global piecewise-smooth model $D(x,y)$ with implied occlusions. Consensus mean $\bar{d}(x,y)$ is initially erroneous in many places but expresses high variance $\sigma^2(x,y)$ there. Variance and error decrease as iterations guide toward the true boundary.
    }
    \label{fig:consensus_viz}
\end{figure*}

It is challenging to implement the alternating approach in a way that succeeds despite the erratic behavior of matching costs $C(x,y,d)$. Noise, textureless surfaces, repetitive textures, and other effects create many local minima and sometimes false global minima, even when matching signals are aggregated over spatial patches of the visual field (e.g., green curves in Fig.~\ref{fig:patch_vote}). Our strategy is to aggregate matching within dense overlapping patches at multiple scales, and to share information among these patches through an evolving family of per-pixel Gaussian disparity distributions that we call the \emph{consensus}. The consensus is visualized in Fig.~\ref{fig:consensus_viz} by its means $\bar{d}(x,y)$ and standard deviations $\sigma(x,y)$. During iterations, the consensus summarizes the disparity information at each pixel from all unoccluded patches that contain the pixel.

Specifically, let $\mathcal{P} = \{ p \}$, a set of densely overlapping patches $p$ of multiple sizes, including a complete subset of patches that each comprise a single pixel $(x,y)$. Equip each patch with an evolving state $\{w_p, d_p,\sigma_p\}$ representing a patch's occlusion status $w_p\in\{0,1\}$ and a \emph{disparity message} $(d_p,\sigma_p)\in\mathbb{R}^2$, visualized by the orange curves in Fig.~\ref{fig:patch_vote}, that represents a balance between the patch's local matching cost and the evolving global model $D(x,y)$. 

The multiscale descent alternates between:
\begin{enumerate}
    \item each patch updates its occlusion status $w_p$ and local disparity message $(d_p,\sigma_p)$ based on the current boundary $\phi(x,y)$ and global models $\mathbf{\Theta}_1, \mathbf{\Theta}_2$;
    
    \item local disparity messages are collected in the consensus $\bar{d}(x,y),\sigma(x,y)$;
    
    \item an update of the global models $\mathbf{\Theta_1},\mathbf{\Theta_2}$ and boundary $\phi(x,y)$ based on the consensus.
\end{enumerate}
The details of each step follow. 

\vspace{1mm}
\noindent \textbf{1. Updating local disparity messages.}
\vspace{1mm}
We initially assume all patches are unoccluded or \emph{valid}: $\forall p, w_p = 1$. At subsequent iterations, each patch updates its validity using 
\begin{multline}
\label{eqn:upward}
\small
  w_p = \max \limits_{(x,y) \in p} \phi(x,y)  > 0  \\
  \oplus \min\limits_{(x,y) \in p} \phi(x+\Delta\theta(x,y;\mathbf{\Theta}_1,\mathbf{\Theta}_2),y)  < 0,
\end{multline}

\noindent where $\oplus$ is the logical XOR operator. This says that a valid patch is neither contained in an occluded regions nor includes portions of both foreground and visible-background.

Each valid patch also updates its local disparity message based on a combination of its local matching cost and the current piecewise-smooth global disparity map:
\begin{equation}\label{eq:local_message}
\small
    d_p = \argmin_{d}  C_p(d)  \quad \text{and} \quad \sigma_p = \frac{d_\text{max}}{\langle C_p(d)\rangle- \min{C_p(d)}},
\end{equation}
where $C_p(d) = \sum_{(x,y) \in p} ( C(x,y,d)  + \beta | d - D(x,y)| )$ with $\beta$ a tunable parameter and $D(x,y)$ computed using Eq.~\ref{eqn:D}. 
Fig.~\ref{fig:patch_vote} shows two examples of the evolving local disparity messages. We interpret $d_p,\sigma_d$ as parameters of an evolving Gaussian-like approximation, and depict them by drawing
\begin{equation}\label{eq:patch-gaussian}
\small
    f_p(d)=\max{C_p(d)}-\left(\max{C_p(d)}-\min{C_p(d)}\right) e^{\frac{-(d-d_p)^2}{2\sigma^2_p}}.
\end{equation}

\vspace{1mm}
\noindent \textbf{2. Updating consensus.}
 The consensus aggregates the local disparity messages from all valid patches and is updated as a product of Gaussians:
 \begin{equation}
 \small
 \begin{split}
        \frac{1}{\sigma^2(x,y)} &= \sum_{\substack{p \ni (x,y) \\  w_p =1}} \frac{1}{\sigma_{p}^{2}}, \quad 
      \overline{d}(x,y) =\Biggl[\sum_{\substack{p \ni (x,y) \\  w_p =1}} \frac{d_{p}}{\sigma_{p}^{2}}\Biggl] \sigma^{2}(x,y).
 \end{split}
 \label{eqn:pixel_consensus_vote}
 \end{equation}

\vspace{1mm}
\noindent \textbf{3. Updating global shapes and boundary.}
\label{sec:global_updates}
The global models $\mathbf{{\Theta}}_{1}$ and $\mathbf{{\Theta}}_{2}$ are updated by maximum likelihood estimation:
\begin{equation}
\label{eqn:global_shape}
\begin{split}
    \mathbf{\Theta}_{j=1,2} & =  \argmin\limits_{\mathbf{\Theta}_j} \sum_{\substack{(x,y) \in \Omega_{j}}}
    \tfrac{(\Theta_j(x,y) - \overline{d}(x,y))^{2}}{2 \sigma^{2}(x,y)},
\end{split}
\end{equation}
with $\Omega_1,\Omega_2$ the foreground and background respectively. This requires solving two linear systems of equations based on the consensus values at many pixels, one for each region $j$: 
\begin{equation}
\begin{split}
        \frac{1}{\sigma^2(x,y)} \mathbf{\Theta}_j\mathbf{U}(x,y) =   \frac{\overline{d}(x,y)}{\sigma^2(x,y)} \text{for } (x,y) \in \Omega_{j},
\end{split}
\end{equation}
 Lastly, to complete the alternation round, the function $\Delta\theta(x,y; \mathbf{\Theta}_1,\mathbf{\Theta}_2)$ is updated using Eq.~\ref{supeqn:delta_theta}, and the boundary $\phi(x,y)$ is updated by descent using a discrete approximation to Eq.~\ref{eqn:dphi_dt}.

\section{Cooperative Architecture}
\label{sec:cooperative_arch}

\begin{figure}[t]
     \centering
     \includegraphics[width=0.49\textwidth]{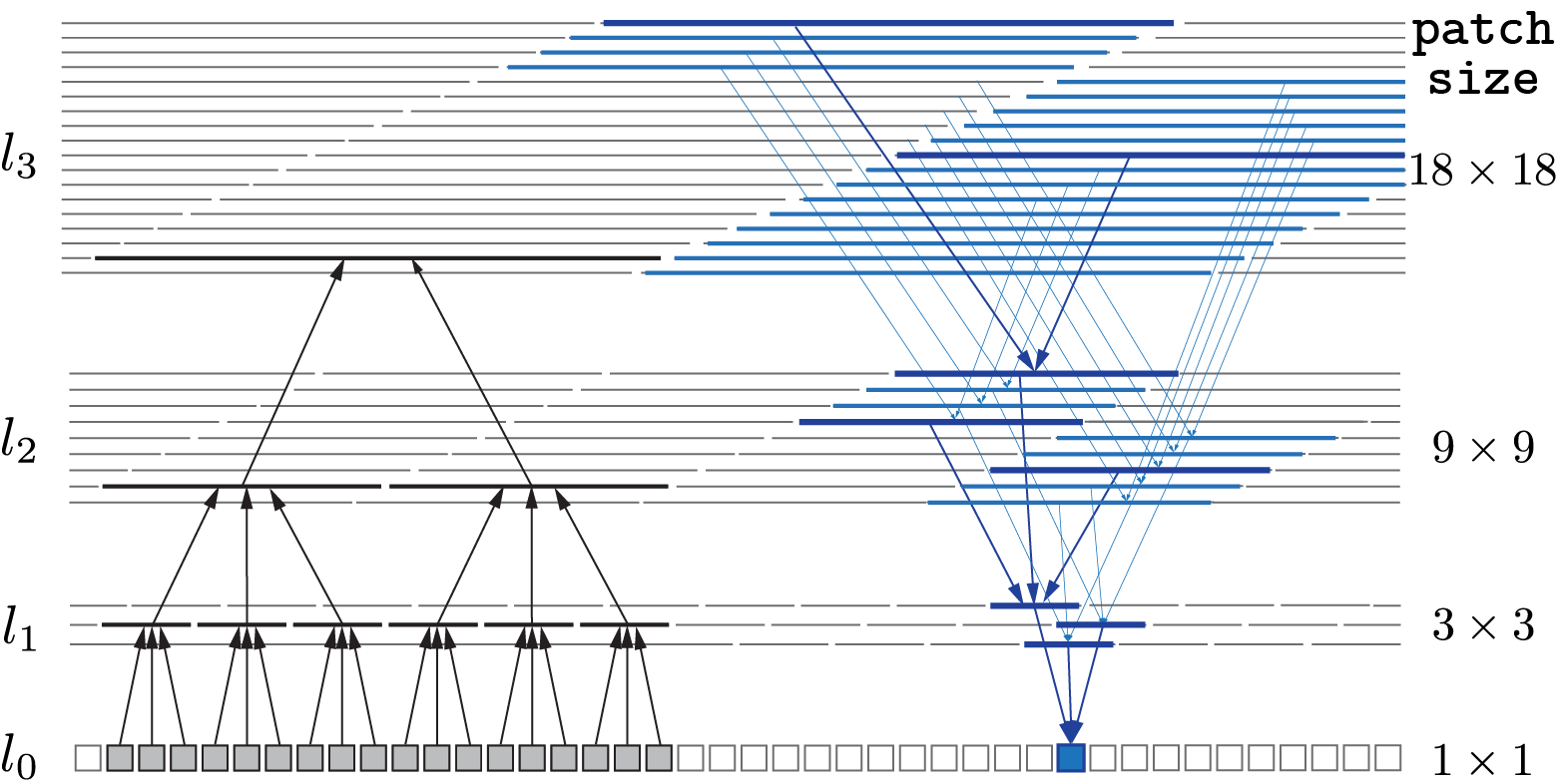}
    \caption{Overlapping multi-scale patches adapted from Chakrabarti et al.~\cite{chakrabarti2015low}, with single pixels in layer $l_0$. Upward pass: Patch aggregates information from pixels it contains via messages from children to parents. Downward pass: Pixel aggregates information from patches that contain it via messages from parents to children.}
         \label{fig:overlapping_patches}
\end{figure}

The multiscale alternating descent can be implemented in a fully-cooperative manner, with messages passing between sparse connections among patches. We describe one such implementation here, based on an undecimated, hierarchical set of multiscale patches as used in~\cite{chakrabarti2015low}. The patches $\mathcal{P} = \{ p \}$ are arranged in a hierarchy as depicted in Fig.~\ref{fig:overlapping_patches} for a 1D epipolar scanline. The single-pixel patches are at the lowest level $l_0$ and each patch $p$ in levels $l_k, k>0$ is a union of its non-overlapping child patches,  denoted $\{p^-\}$. For example, a $9 \times 9$ patch has nine $3 \times 3$ children (only three of which are shown in the 1D depiction). Each patch below the top layer shares a bi-directional connection to its parent patches, denoted $\{p^+\}$. Each $l_0$-patch also connects laterally to its $\pm d_{\max}$ neighbors along its epipolar scanline. We use notation $|p|$ to denote the size of patch $p$.


We associate with each patch $p$ a computational unit as well as read-only memory storing its location $(x,y)$ and the local portion of the basis 
$\{\mathbf{U}(x,y),  (x,y)\in p\} $ as a $|p| \times m$ matrix $\mathbf{U}_p$
and writable memory for the local matching cost and a local state. Pixel-level patches $p\in l_0$ are indexed by $(x,y)$, and each one stores $\{C(x,y,d), d \in [0,d_{\max}]\}$ and an evolving state comprising $3+2m$ values: $\{\bar{d}(x,y); \sigma(x,y); \phi(x,y); \mathbf{\Theta}_1; \mathbf{\Theta}_2\}$. Higher-level patches $p$ store evolving states $\{w_p, d_p,\sigma_p\}\in\{0,1\}\times\mathbb{R}^2$ representing a patch's validity and local disparity message. The cooperative process proceeds as follows.


\begin{figure*}[t]
    \centering
    \includegraphics[width=0.75\textwidth]{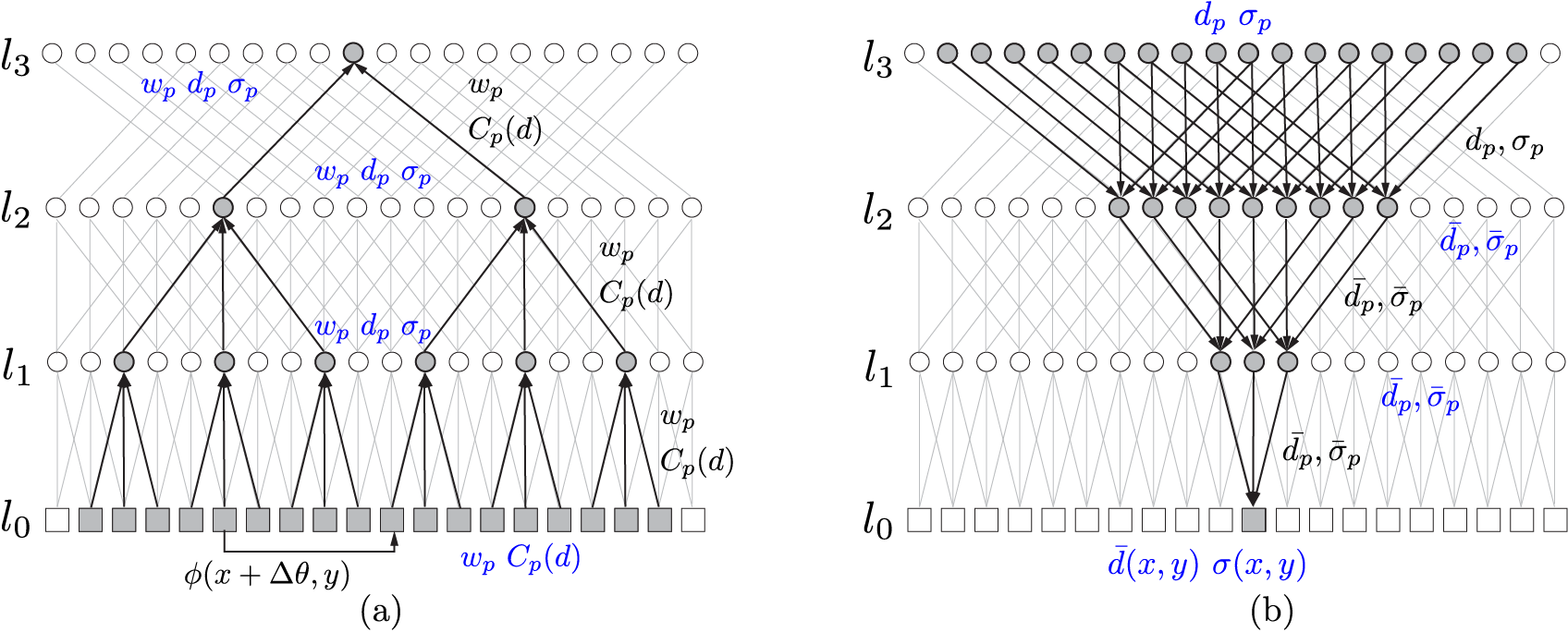}
    \caption{Hierarchical patches from Fig.~\ref{fig:overlapping_patches} depicted as a network, with one node per patch and edges between parents and children. Blue quantities are computed locally at a node and black quantities are messages passed along edges. (a) Upward pass: Each parent patch updates its valididity $w_p$ and local disparity message $d_p$ and $\sigma_p$ using all of the pixels it contains. (b) Downward pass: Each pixel updates its consensus values $\overline{d}(x,y)$ and $\sigma(x,y)$ by collecting input from all valid patches that contain it.}
    \label{supfig:algo_upward_downward}
\end{figure*}
\begin{figure*}[t!]
    \centering
    \includegraphics[width=0.63\textwidth]{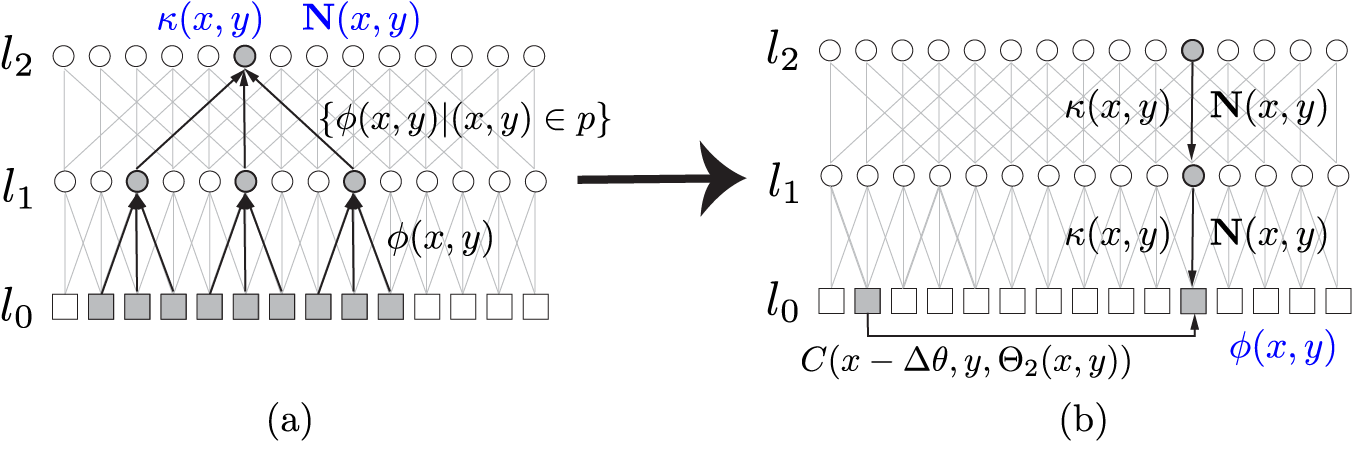}
    \caption{Boundary update. (a) Computing curvatures $\kappa(x,y)$ and normals $\mathbf{N}(x,y)$. Their values are weighted averages of $\phi(x,y)$ passed up from children to parents. (b) Each patch $p\in l_0$ updates $\phi(x,y)$ using lateral connections of radius at most $d_\text{max}$, where $\Delta\theta$ is computed using Eq.~\ref{supeqn:delta_theta}.}
    \label{supfig:algo_phi}
\end{figure*}

\vspace{1mm}
\noindent \textbf{1. Upward pass: Updating local disparity messages.}
Patch validities and disparity messages are updated in an upward pass, with messages passing from children to parents, as depicted in black in Fig.~\ref{fig:overlapping_patches} and in Fig.~\ref{supfig:algo_upward_downward}(a). For validities, each patch $p$ computes an internal pair of binary values $W_p = \{f_p,b_p\}$ indicating, respectively, whether it contains a foreground pixel and a visible-background pixel. With this notation, Eq.~\ref{eqn:upward} becomes $w_p = f_p \oplus b_p$,
and computing $W_p = \{f_p,b_p\}$ for $p\in l_0$ is trivial using lateral connections. For patches in higher levels, $p\in l_k, k>0$, we sequentially compute
\begin{equation}
    f_p=\bigvee_{q\in\{p^-\}} f_{q}, \quad b_p=\bigvee_{q\in\{p^-\}} b_q,
\end{equation}
with $\bigvee$ the logical OR operator.

Local disparity messages are updated by first computing
\begin{equation}
    C_p(d) = C(x,y,d)  + \beta \big| d - D(x,y) \big| 
\end{equation}
at patches $p\in l_0$ and then sequentially computing 
\begin{equation}
    C_p(d) = \sum_{q \in \{p^{-}\}} C_{q}(d)
\end{equation}
for $p\in l_k, k>0$.

\vspace{1mm}
\noindent \textbf{2. Downward pass: Updating consensus.}
Consensus values $\bar{d}(x,y),\sigma(x,y)$ are updated using messages from parents to children in a downward pass, as depicted in blue in Fig.~\ref{fig:overlapping_patches} and in Fig.~\ref{supfig:algo_upward_downward}(b). Each patch computes
\begin{equation}
\begin{split}
  \label{supeqn:d_p}
    \frac{1}{\overline{\sigma}_p^2} & = w_p \frac{1}{\sigma_p^{2}} + \sum_{\substack{q \in \{p^+\}}} \frac{1}{\overline{\sigma}_{q}^{2}} \\
    \overline{d}_{p} & = w_p \frac{d_p}{\sigma_p^2} + \sum_{\substack{q \in \{p^+\}}} \frac{\overline{d}_{q}}{\overline{\sigma}_{q}^2}  
\end{split}
\end{equation}
and passes $\overline{d}_p$ and $\overline{\sigma}_p$ to its children. The consensus at each pixel $(x,y)$ is the result at the pixel level:
\begin{equation}
    \sigma(x,y)=\overline{\sigma}_p \quad \text{and} \quad \overline{d}(x,y)=\overline{d}_p \sigma^2(x,y),\quad p\in l_0.
\end{equation}

\vspace{1mm}
\noindent \textbf{3. Updating global shapes and boundary.}
Global shapes $\mathbf{\Theta}_{j=1,2}$ can be computed cooperatively in $l_0$ by first computing $m \times m$ local correlation matrix
\begin{equation}
\begin{split}
    \mathbf{A}_p = \frac{1}{\sigma^4(x,y)} \mathbf{U}_p^T \mathbf{U}_p.
\end{split}
\end{equation}
and cross-correlation $m$-vector 
\begin{equation}
    b_p = \mathbf{U}_p^T c_p
\end{equation}
with $|p|$-vector $c_p = \frac{\bar{d}(x,y)}{\sigma^4(x,y)}, (x,y) \in p$ in each patch separately, and then updating using Tron and Vidal's consensus averaging algorithm~\cite{tron2011distributed} with the lateral connections in $l_0$.




Finally, boundaries are updated via Eq.~\ref{eqn:dphi_dt} in an upward pass using per-pixel curvature $\kappa(x,y)$ and normal $\mathbf{N}(x,y)$ values given by local filters in $\ell_2$, as depicted in  Fig.~\ref{supfig:algo_phi}. Then $\phi(x,y)$ is updated using lateral connections between $l_0$ patches that share $C(x-\Delta\theta(x,y,\mathbf{\Theta}_1, \mathbf{\Theta}_2),y,\Theta_2(x,y))$. 
Lastly, $\Delta\theta(x,y, \mathbf{\Theta}_1, \mathbf{\Theta}_2)$ is updated using Eq.~\ref{supeqn:delta_theta}.

\section{Experiments}
\begin{figure*}[t!]
    \centering
    \includegraphics[width=\textwidth]{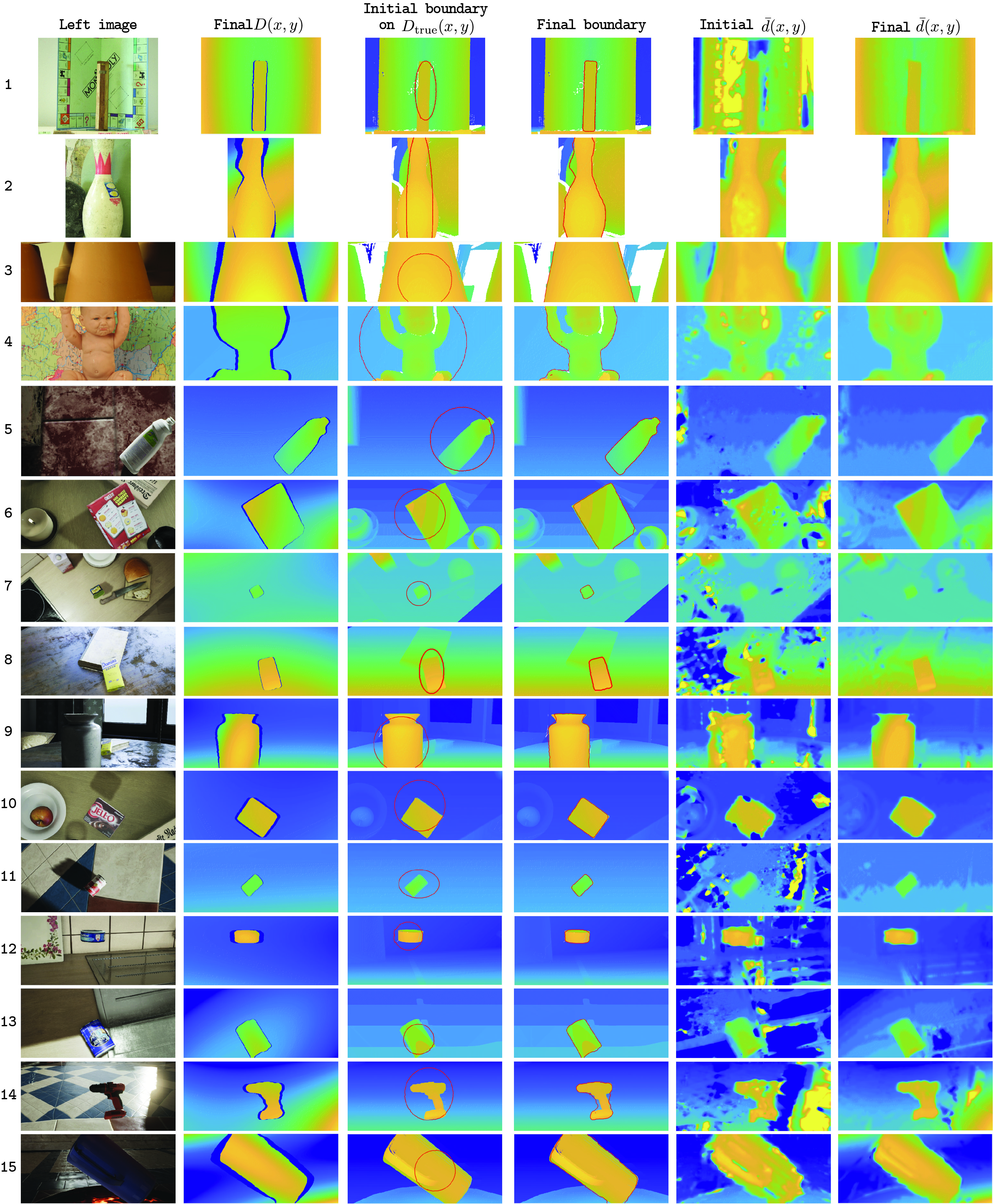}
    \caption{\textbf{Qualitative results on two-layer scenes.} From left to right: stereo left image; final disparity map $D(x,y)$ with occlusions shown in deep blue; initial boundary superimposed on ground-truth disparity map (white is missing data in the ground truth); converged boundary; consensus $\bar{d}(x,y)$ at initialization; and converged consensus $\bar{d}(x,y)$. Columns  5 and 6 show that the model succeeds in recovering from noisy matching costs, as are evident in the erroneous consensus values $\bar{d}(x,y)$ at initialization.}
    \label{fig:all_results}
\end{figure*}

\begin{figure*}[t!]
    \centering
    \includegraphics[width=0.8\textwidth]{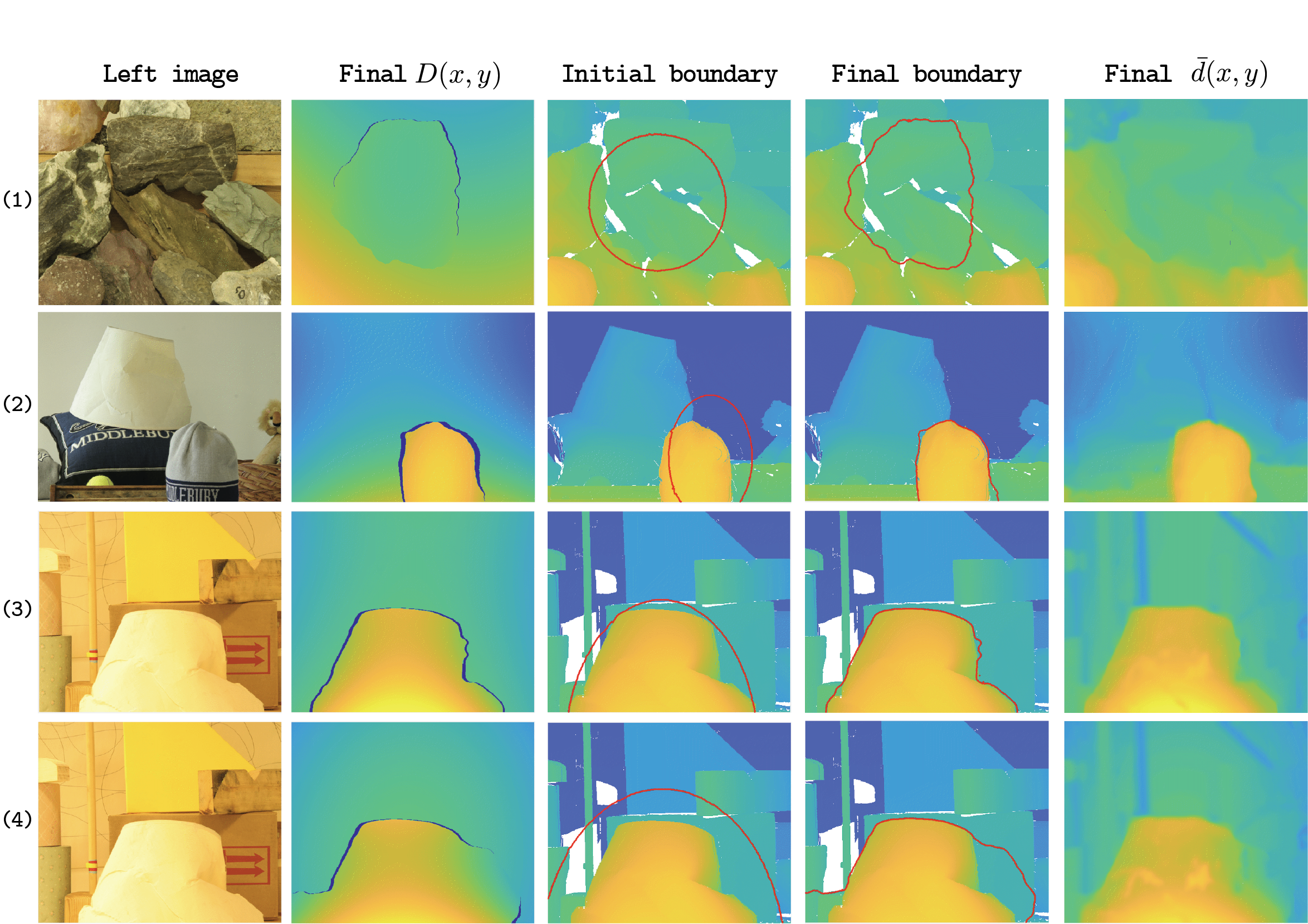}
    \caption{\textbf{Qualitative results on more complex scenes.} From left to right: stereo left image; final disparity map $D(x,y)$ with occlusions shown in deep blue; initial boundary superimposed on the ground-truth disparity map (white is missing data in the ground truth); final boundary; and converged consensus $\bar{d}(x,y)$.}
    \label{fig:not2planes}
\end{figure*}

\vspace{1mm}
\noindent \textbf{Datasets.} 
Existing stereo benchmarks do not include two-layer, figure-ground scenes, so we curate our own using four photographic crops from Middlebury 2006~\cite{hirschmuller2007evaluation} and eleven renderings from Falling Things~\cite{tremblay2018falling}, shown in Fig.~\ref{fig:all_results}. The latter includes ground-truth foreground boundaries, and we manually annotate them in the former. We additional use three multi-layer scenes from Middlebury 2006~\cite{hirschmuller2007evaluation} to examine the model's behavior when a scene differs significantly from a two-layer one (Fig.~\ref{fig:not2planes}).

\vspace{1mm}
\noindent \textbf{Matching and boundary signals.} Our model can use any underlying matching and boundary signals, including those from learned, deep models (e.g.,~\cite{wang2019local,vzbontar2016stereo,xie2015holistically}). Here we choose to use simpler, weaker signals to focus the burden on our optimization and to ensure we are providing a conservative estimate of achievable performance. 

We use absolute difference of intensity for matching, $C(x,y,d) = \sum_c |I_l(x+d,y,c) - I_r(x-d,y,c)|$. We generate the monocular boundary cost $B_{\text{m}}(x,y,d)$ from Eq.~\ref{eq:monocular-boundary} using edge maps created by convolving a $3 \times 3$ Sobel filter with each input image, thresholding the responses, and converting the resulting binary edge maps to distance functions $E_l(x',y')$ and $E_r(x',y')$ where $E=0$ at detected edges and $E>0$ away from edges. For the occlusion boundary cost $B_{\text{o}}(x,y,d)$, we compute the epipolar gradient magnitude $|\partial C(x,y,d)/\partial x|$, threshold the result, and convert the binary volumetric function to a distance function where $B_{\text{o}} = 0$ at the detected boundaries. We linearly normalize the values in each of $C$, $B_{\text{m}}$ and $B_{\text{o}}$ to be in the range $[0,1]$. The thresholds for monocular edge maps and occlusion boundary maps vary slightly from scene to scene and are optimized for each scene.

\begin{figure}[t]
    \centering
    \includegraphics[width=0.47\textwidth]{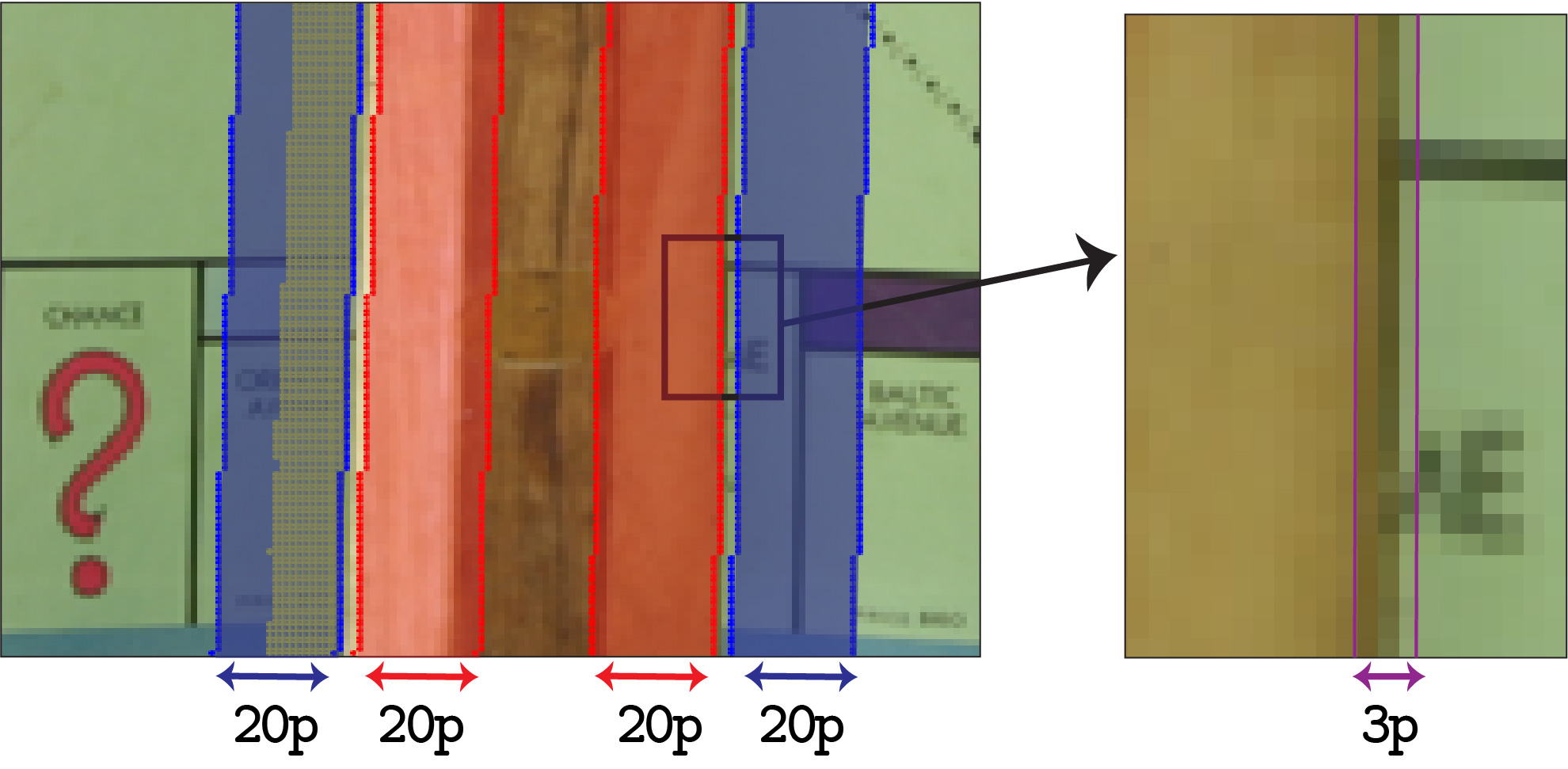}
    \caption{Evaluation area. We measure accuracy in regions $\pm 20$ pixels in the epipolar direction from true boundaries, excluding $\pm1$ pixels which are often affected by lens blur. In this example, red shows evaluated foreground regions, blue shows evaluated background regions, and yellow dots are ground-truth occluded pixels.}
    \label{fig:eval_area}
\end{figure}

\begin{table*}[t!]
\centering
\setlength{\tabcolsep}{5.6pt}
\begin{tabular}{lcccccccccccccccccc}
\toprule
Image & 1 & 2 & 3 & 4 & 5 & 6 & 7 & 8 & 9 & 10 & 11 & 12 & 13 & 14 & 15 & & Avg occlusion F1 \\
\midrule 
SGM~\cite{hirschmuller2009matching} & 0.39 & 0.38 & 0.42 & 0.65 & 0.23 & 0.32 & 0.00 & 0.02 & 0.58 & 0.67 & 0.04 & 0.87 & 0.27 & 0.61 & 0.72 & & 0.41 \\
BM-LR & 0.37 & 0.34 & 0.48 & 0.70 & 0.01 & 0.22 & 0.00 & 0.00 & 0.57 & 0.72 & 0.18 & 0.92 & 0.07 & 0.59 & 0.44 & & 0.37 \\
KZ~\cite{kolmogorov2001computing} & 0.14 & 0.55 & 0.76 & \textbf{0.78} & 0.76 & 0.53 & \textbf{0.68} & 0.70 & \textbf{0.83} & 0.82 & 0.57 & 0.90 & 0.46 & 0.60 & 0.85 & & 0.66 \\
HSM~\cite{yang2019hierarchical} & 0.55 & \textbf{0.63} & 0.98  & 0.62 & 0.49 & 0.31 & 0.04 & 0.60 & 0.64 & 0.83 & 0.18 & 0.87 & 0.23 & 0.70 & \textbf{0.90} & & 0.57 \\
Ours & \textbf{0.87} & 0.60 & \textbf{0.99} & \textbf{0.78} & \textbf{0.86} & \textbf{0.55} & 0.61 & \textbf{0.75} & 0.75 & \textbf{0.98} & \textbf{0.64} & \textbf{0.95} & \textbf{0.79} & \textbf{0.81} & 0.88 & & \textbf{0.79} \\
\bottomrule
\end{tabular} 
\caption{Boundary accuracy measured by occlusion F1-score for the fifteen scenes from Fig.~\ref{fig:all_results}, as well as test set average. 
}
\label{tbl:result}
\end{table*}

\begin{table*}[t!]
\centering
\setlength{\tabcolsep}{4.5pt}
\begin{tabular}{lccccccccccccccccc}
\toprule
Image & 1 & 2 & 3 & 4 & 5 & 6 & 7 & 8 & 9 & 10 & 11 & 12 & 13 & 14 & 15 & & Avg bad-4.0 \\
\midrule
SGM~\cite{hirschmuller2009matching} & 13.10 & 21.79 & 43.66 & 10.30 & 10.65 & 29.53 & 8.09 & 11.46 & 17.41 & 8.94 & 8.03 & 19.75 & 14.27 & 14.17 & 32.38 &  & 17.57 \\
BM-LR & 19.88 & 30.71 & 36.97 & 9.21 & 12.78 & 32.36 & 13.42 & 13.13 & 22.07 & 15.47 & 7.83 & 16.42 & 17.09 & 17.86 & 45.06 & & 20.68 \\
KZ~\cite{kolmogorov2001computing} & 31.26 & 13.15 & 49.97 & 3.84 & 3.53 & 18.86 & 2.10 & 3.22 & 12.52 & 1.62 & 2.57 & 9.99 & 24.60 & 57.40 & \textbf{29.05} & & 17.58 \\
HSM~\cite{yang2019hierarchical} & 1.69 & \textbf{3.35} & \textbf{0.25} & \textbf{1.80} & 0.99 & \textbf{2.92} & \textbf{0.00} & \textbf{0.00} & \textbf{2.59} & \textbf{0.89} & \textbf{0.00} & \textbf{5.69} & \textbf{2.38} & \textbf{3.18} & 63.08 & & \textbf{5.92} \\
Ours & \textbf{0.71} & 24.84 & 7.54 & 8.05 & \textbf{0.17} & 30.89 & \textbf{0.00} & 1.50 & 41.38 & 1.73 & 0.60 & 10.23 & 3.20 & 32.34 & 77.81 & & 16.07 \\
\bottomrule
\end{tabular}
\caption{Disparity accuracy near boundaries, as measured by bad-4.0, for the fifteen scenes from Fig.~\ref{fig:all_results}, along with test set average. Deep model HSM outperforms other methods, and our model performs comparable to or better than the others. Note that disparity errors are  higher overall for scenes that are closer to the camera (e.g., scene 15).
}
\label{tbl:add_result}
\end{table*}

\vspace{1mm}
\noindent \textbf{Metrics and comparison methods.}  We measure boundary accuracy using precision and recall---summarized by F1 score---of estimated binary occlusion labels in a region of $\pm 20$ epipolar pixels around the ground-truth boundary (see Fig.~\ref{fig:eval_area}). As a secondary comparison, we measure accuracy of estimated disparities in the true, mutually-visible subset of this region, using the fraction of pixels at which the disparity error is greater than four pixels (bad-4.0).

For comparisons, we consider four categories of techniques and select one representative method from each of them: 
\begin{enumerate}
    \item BM-LR: Block matching and pixel-wise winner-take-all, followed by a left-right consistency check (e.g.,~\cite{fua1993parallel});
    \item SGM: Semi-global matching, based on multi-directional 1D dynamic programming~\cite{hirschmuller2009matching};
    \item KZ: Graph-cuts with occlusion~\cite{kolmogorov2001computing};
    \item HSM: A deep CNN model that augments training with fake occlusions~\cite{yang2019hierarchical}.
\end{enumerate}

\vspace{1mm}
\noindent \textbf{Implementation details.} For each input, we manually initialize our model to an elliptical boundary as shown in Figures~\ref{fig:all_results} and~\ref{fig:not2planes}. We re-initialize $\phi(x,y)$ to a signed distance function every $10$ iterations, and we apply a $7\times7$ median filter to $\phi(x,y)$ after every iteration to eliminate thin structures. Parameters values are: $dt= 0.2$; $\alpha_1= 0.2$; $\alpha_2= 0.8$; $\alpha_3= 0.1$; $\mu=4.0$ and $\beta= 0.4/d_\text{max}$. Our implementation of the alternating descent algorithm and data are open sourced at \url{https://github.com/jialiangw/levelsetstereo}.

We use MATLAB 2019b implementations of SGM and BM-LR and the authors' implementation for KZ and HSM. We use the HSM authors' weights, trained on a mixture of datasets including Middlebury, and the occlusion geometry discussed in Sec.~\ref{sec:45degree} to get occlusions.


\begin{figure*}[t!]
    \centering
    \includegraphics[width=0.95\textwidth]{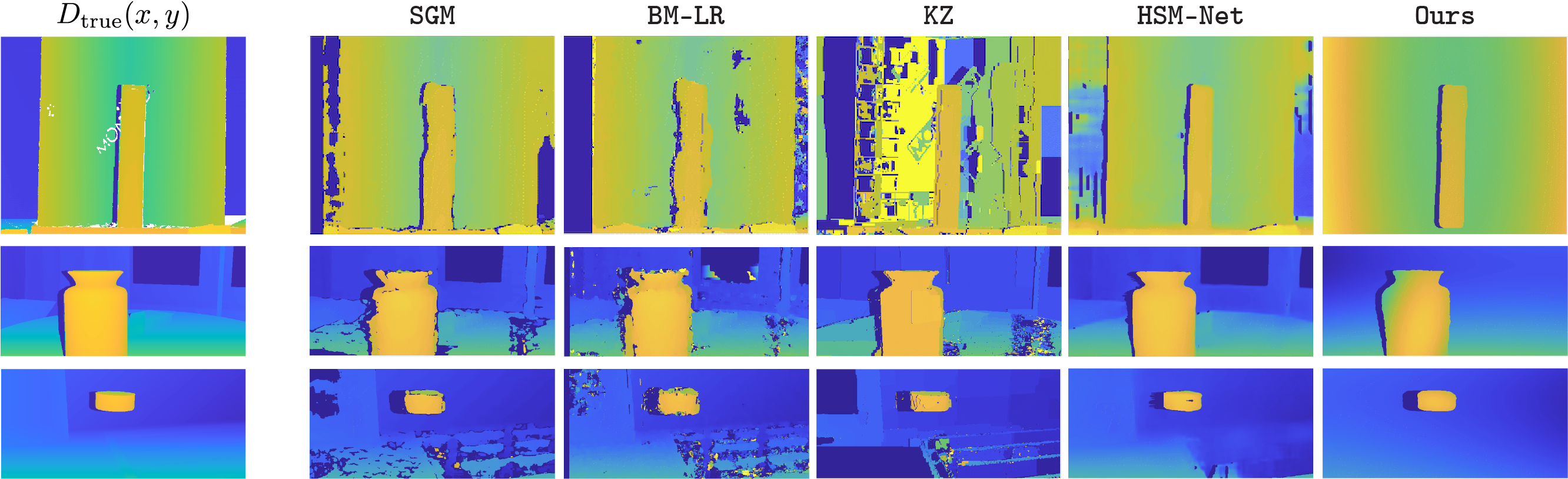}
    \caption{\textbf{Comparison to existing methods.}  Our model's final disparity and occlusion maps compared to the ground truth and to outputs of other methods. Occlusions are shown in deep blue, and white in column 1 is missing data in the ground truth.}
    \label{fig:comparison}
\end{figure*}

\subsection{Results on two-layer scenes}

Our qualitative results on two-layer, figure-ground scenes are shown in Fig.~\ref{fig:all_results}, and some comparisons with other methods are shown in Fig.~\ref{fig:comparison}. The quantitative results are summarized in Tables~\ref{tbl:result} and~\ref{tbl:add_result}. Our model provides the highest boundary accuracy (lowest occlusion F1-score) in all but four scenes, and its disparity error near the boundaries (bad-4.0 error) is better than most. The deep model, HSM, achieves substantially lower disparity error as expected, but its occlusion accuracy is middling. This supports the belief that iterative bottom-up models like ours could complement disparity estimates from deep feed-forward models. 


Our model's least accurate boundaries, meaning those with lowest occlusion F1-score, occur in scenes 2, 6, 7, 9 and 11. The qualitative results in Fig.~\ref{fig:all_results} show that this typically occurs when the foreground or background shape deviates substantially from a quadratic model, as can be seen in scenes 2, 6 and 9. F1-scores in scenes 7 and 11 are low for a different reason: the ground truth contains very few occluded pixels, making F1-score very sensitive to boundary misalignment. Our model also performs poorly in scene 4, where the background regions between the doll's head and arms are almost entirely occluded or have weak matching cues (especially with our simple matching cost), causing our model not to properly group these regions with the rest of the background.

\subsection{Results on more complex scenes}

Fig.~\ref{fig:not2planes} qualitatively demonstrates how the model behaves for three scenes that deviate substantially from two-layer. Note that rows (3) and (4) depict results for the same scene but with different initializations. In scene (1) there are few boundary occlusion cues, so $B_o$ is roughly constant and the model fits its two quadratic shapes to match monocular color boundaries and minimize matching cost. The predicted boundaries are not meaningful, but the disparity map still provides a coarsened approximation to the ground truth. In scenes (2) and (3), the model's foreground boundary is more useful but still inaccurate. Row (4) shows that when a scene deviates substantially from two quadratic layers, our method can be sensitive to local minima caused by the boundary cues induced by secondary objects. Here we initialize the boundary to a larger ellipse and find that the model converges to a local minimum caused by the painted color boundaries on the box behind the lampshade. Nonetheless, the converged consensus $\bar{d}(x,y)$ and the final disparity map $D(x,y)$ still provide a coarsened approximation to the ground truth. 





\section{Conclusion}
By exploiting occlusion and matching cues on equal footing, our model localizes foreground contours in a variety of bi-layer scenes. The algorithm can be implemented in a cooperative manner, with messages passing along sparse connections between units with different receptive field sizes. It provides promising results despite having only five tunable parameters and using local matching and boundary signals that are as simple as possible.

An algorithm like ours can be used as-is for scenes with isolated foreground objects and for tasks, like grasping, that rely heavily on having an accurate foreground mask. Broadening to general scenes and tasks will require upgrading the underlying matching and boundary signals to more sophisticated (e.g.,~``deep''~\cite{vzbontar2016stereo,wang2019local,xie2015holistically}) alternatives, and perhaps tuning the parameters jointly by unrolling our model's iterations in time. It could also be combined with a secondary processing step to recover higher-fidelity disparity details within the two smooth regions (i.e., surface $+$ parallax). 

Open question include how to increase the number of regions beyond two (e.g., using multi-phase level sets~\cite{vese2002multiphase}), how to reduce sensitivity with respect to initialization, and how to combine this sort of bottom-up stereo boundary processing with complementary top-down information from deep feed-forward models.
Another direction is to explore ways of encoding geometric relationships between smooth layers, such as when planar facets join at a crease. How and whether such relationships are encoded in the human visual system remains a mystery (see~\cite{nakayama1992experiencing,ehrenstein1998early}) that computational models may help solve.


%
\IEEEpeerreviewmaketitle

\appendices  

\section{Derivation of update to $\phi(x,y)$} 

Fix global models $\mathbf{\Theta}_1$ and $\mathbf{\Theta}_2$. 
Rewrite energy function, Equation~\ref{eqn:loss_function}, as
\begin{equation}
\begin{split}
    J(x, \phi, \phi_{+}, \nabla\phi) = & \int_\Omega F(x,\phi, \phi_{+}, \nabla\phi) dx,
\end{split}
\end{equation}
with
\begin{equation*}
    \begin{split}
        & F(x,\phi, \phi_{+}, \nabla\phi) =  F_C(x,\phi,\phi_{+}) +  F_B(x,\phi , \nabla\phi), \\
        & F_C(x,\phi,\phi_{+}) =  H(\phi) C_{\Theta_1} + (1-H(\phi))(1-H(\phi_{+})C_{\Theta_2}, \\
        & F_B(x,\phi , \nabla\phi) =  \mu B_{\Theta_1} \delta(\phi) |\nabla \phi|.
    \end{split}
\end{equation*}

The Euler-Langrange equations are
\begin{equation}
\footnotesize
\label{supeqn:first_EL}
        \frac{\partial J(x, \phi, \phi_{+}, \nabla\phi)}{\partial \phi} =  \frac{\partial F(x, \phi, \phi_{+}, \nabla\phi)}{\partial \phi} - \frac{d}{dx}\biggl( \frac{\partial F(x, \phi, \phi_{+}, \nabla\phi)}{\partial \nabla \phi} \biggl)
\end{equation}
\begin{equation}
\label{supeqn:second_EL}
\footnotesize
        \frac{\partial J(x, \phi, \phi_{+}, \nabla\phi)}{\partial \phi_{+}}  =  \frac{\partial F(x, \phi, \phi_{+}, \nabla\phi)}{\partial \phi_{+}} - \frac{d}{dx}\biggl( \frac{\partial F(x, \phi, \phi_{+}, \nabla\phi)}{\partial \nabla \phi_{+}} \biggl)   .  \\
\end{equation}

The first equation~(\ref{supeqn:first_EL}) has two parts. The first part is:
\begin{equation*}
\small
    \begin{split}
        \frac{\partial F(x, \phi, \phi_{+}, \nabla\phi)}{\partial \phi} & = \frac{\partial F_C(x,\phi,\phi_{+})}{\partial \phi} + \frac{\partial F_B(x,\phi , \nabla\phi)}{\partial \phi}, \\
        \text{where } \quad    \frac{\partial F_B(x,\phi , \nabla\phi)}{\partial \phi} & = \mu
        \frac{\partial}{\partial \phi} B_{\Theta_1} \delta(\phi) |\nabla \phi| \\
        & = \mu B_{\Theta_1} |\nabla \phi| \frac{\partial}{\partial \phi} \delta(\phi),\\
    \end{split}
\end{equation*}
\begin{equation*}
\small
    \begin{split}
        \text{and } \quad &  \frac{\partial F_C(x,\phi,\phi_{+})}{\partial \phi}  \\
        = \quad &  \frac{\partial}{\partial \phi} H(\phi)C_{\Theta_1}  +  \frac{\partial}{\partial \phi} (1-H(\phi)) (1 - H(\phi_{+})) C_{\Theta_2}   \\
    = \quad &  \delta(\phi) C_{\Theta_1} - \delta(\phi) (1 - H(\phi_{+})) C_{\Theta_2} \\ 
    = \quad & \delta(\phi) C_{\Theta_1}. \\
        \end{split}
\end{equation*}
\begin{figure}[h]
    \centering
    \includegraphics[width=0.45\textwidth]{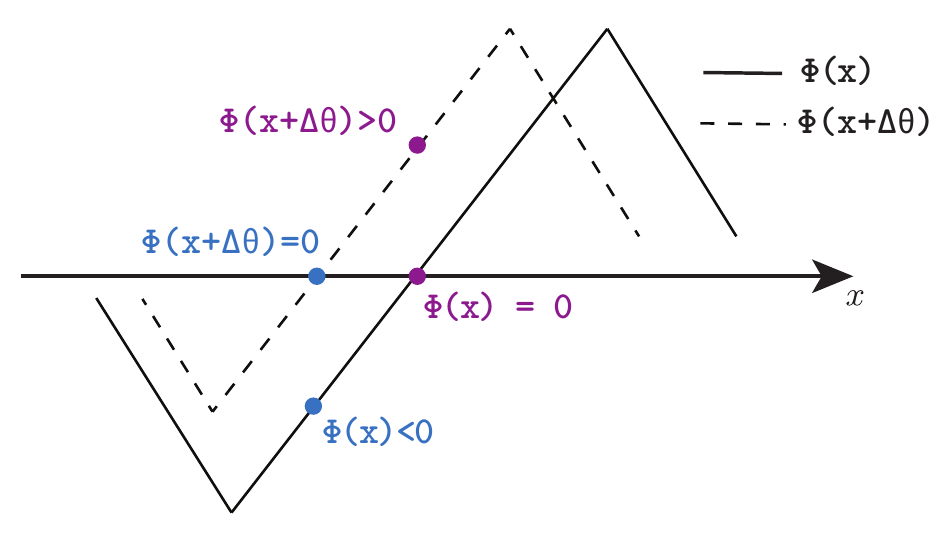}
    \caption{With a slight abuse of notation, assuming there is no background or foreground structure with width less than $\Delta\theta$, it follows that $\phi(x)=0$ implies $\phi(x+\Delta\theta)>0$ (purple points), and that $\phi(x+\Delta\theta)=0$ implies $\phi(x)<0$ (blue points). }
    \label{fig:viz_dphi_dt}
\end{figure}

In the last line, we use the property $1 - H(\phi_{+})=0$ when $\phi = 0$. This assumes that there are no thin foreground structures that violate the ordering constraints~\cite{yuille1984generalized}, as illustrated using the purple points in Figure~\ref{fig:viz_dphi_dt}. Therefore,
\begin{equation}
\label{supeqn:first_part_first_EL}
    \frac{\partial F(x, \phi, \phi_{+}, \nabla\phi)}{\partial \phi} = \delta(\phi) C_{\Theta_1} + \mu  B_{\Theta_1} |\nabla \phi| \frac{\partial}{\partial \phi} \delta(\phi) .
\end{equation}

The second part of Equation~\ref{supeqn:first_EL} is
\begin{equation}
\small
\label{supeqn:second_part_first_EL}
\begin{split}
        & \quad \frac{d}{dx}\biggl( \frac{\partial F(x, \phi, \phi_{+}, \nabla\phi)}{\partial \nabla \phi} \biggl) \\
        & = \frac{d}{dx}\biggl( \frac{\partial F_B(x, \phi, \nabla\phi)}{\partial \nabla \phi} \biggl) \\ 
        & =  \mu \frac{d}{dx}\biggl( \frac{\partial}{\partial \nabla \phi} B_{\Theta_1} \delta(\phi) |\nabla \phi| \biggl) \\
        & =  \mu \delta(\phi) \frac{d}{dx}\biggl( B_{\Theta_1} \frac{\nabla \phi}{|\nabla \phi|} \biggl) + \mu  B_{\Theta_1} \frac{\nabla \phi}{|\nabla \phi|} \frac{d}{dx} \delta(\phi)\\
        & = \mu \delta(\phi) \biggl( B_{\Theta_1} \text{div} \biggl(\frac{\nabla \phi}{|\nabla \phi|}\biggl) + \frac{\nabla \phi}{|\nabla \phi|} \cdot \nabla B_{\Theta_1} \biggl) + \mu  B_{\Theta_1} |\nabla \phi| \frac{\partial \delta(\phi)}{\partial \phi}. \\
\end{split}
\end{equation}
Combining Equations~(\ref{supeqn:first_part_first_EL}) and (\ref{supeqn:second_part_first_EL}) gives
\begin{equation}
\small
\label{supeqn:first_EL_sln}
\begin{split}
    & \quad \frac{\partial J(x, \phi, \phi_{+}, \nabla\phi)}{\partial \phi} \\ 
    & =  \frac{\partial F(x, \phi, \phi_{+}, \nabla\phi)}{\partial \phi} - \frac{d}{dx}\biggl( \frac{\partial F(x, \phi, \phi_{+}, \nabla\phi)}{\partial \nabla \phi} \biggl)  \\ &= \delta(\phi) C_{\Theta_1} - \mu \delta(\phi) \biggl( B_{\Theta_1} \text{div} \biggl(\frac{\nabla \phi}{|\nabla \phi|}\biggl) + \frac{\nabla \phi}{|\nabla \phi|} \cdot \nabla B_{\Theta_1} \biggl).
\end{split}
\end{equation}
The second Euler-Lagrange equation~(\ref{supeqn:second_EL}) also has two parts but the second one is zero, so the right side of that equation is
\begin{equation}
\small
\label{supeqn:second_EL_sln}
    \begin{split}
    & \quad \frac{\partial F(x, \phi, \phi_{+}, \nabla\phi)}{\partial \phi_{+}} \\
    & = \frac{\partial F_C(x,\phi,\phi_{+})}{\partial \phi_{+}} \\
    & = \frac{\partial}{\partial \phi_{+}} H(\phi)M_{\Theta_1} + \frac{\partial}{\partial \phi_{+}} (1-H(\phi)) M_{\Theta_2} (1 - H(\phi_{+}))  \\
         & = - \delta(\phi_{+})(1-H(\phi)) M_{\Theta_2}  = 
         - \delta(\phi_{+}) M_{\Theta_2}, \\ 
    \end{split}
\end{equation}
where we assume there are no thin background structures such that the entire background segment is occluded, as depicted by blue points in Figure~\ref{fig:viz_dphi_dt}. 

Let $x' = x + \Delta\theta(x,y;\mathbf{\Theta}_1,\mathbf{\Theta}_2)$ and sum Equations~(\ref{supeqn:first_EL_sln}) and (\ref{supeqn:second_EL_sln}). We obtain:
\begin{equation}
\small
\label{supeqn:dphi_dt}
\begin{split}
    \frac{d\phi}{dt} & = - \frac{\partial J(x, \phi, \phi_{+}, \nabla\phi)}{\partial \phi} \\
    & = \delta(\phi(x,y)) \biggl[  -  C(x,y, {\Theta}_1(x,y)) \\ 
    + & C(x-\Delta\theta(x,y;\mathbf{\Theta}_1,\mathbf{\Theta}_2),y,{\Theta}_2(x,y)) \\
    + & \mu \Big( B(x,y,{\Theta}_1(x,y)) \kappa(x,y) +  \mathbf{N}(x,y) \cdot \nabla B(x,y,{\Theta}_1(x,y)) \Big) \biggl],
\end{split}
\end{equation}
where 
\begin{equation*}
\small
\kappa(x,y)=\text{div} \left(\frac{\nabla \phi(x,y)}{|\nabla \phi(x,y)|}\right) \quad \text{and} \quad \mathbf{N}(x,y) = \frac{\nabla \phi(x,y)}{|\nabla \phi(x,y)|}    
\end{equation*}
are the curvature and normal of the foreground boundary contour. 

Near the border of the cyclopean visual field $\partial\Omega$, it is possible that $(x-\Delta\theta(x,y;\mathbf{\Theta}_1,\mathbf{\Theta}_2),y)$ is outside of the visual field. When this occurs, we extrapolate the value of $C(x,y,d)$ from the border for use in Equation~\ref{supeqn:dphi_dt}.


%





\ifCLASSOPTIONcaptionsoff
  \newpage
\fi



%

\bibliographystyle{ieeetran}
\bibliography{refs}


%








\end{document}


\onecolumn

{\centering
{\large \textbf{\uppercase{Supplementary Materials: Level Set Stereo for Cooperative Grouping with Occlusion}}\par}
\vspace{0.5cm}
{\large \textit{Jialiang Wang and Todd Zickler}\par}
\vspace{0.2cm}
{\large Harvard University\par}
}

\vspace{1cm}



\begin{figure}[h]
    \centering
    \includegraphics[width=0.65\textwidth]{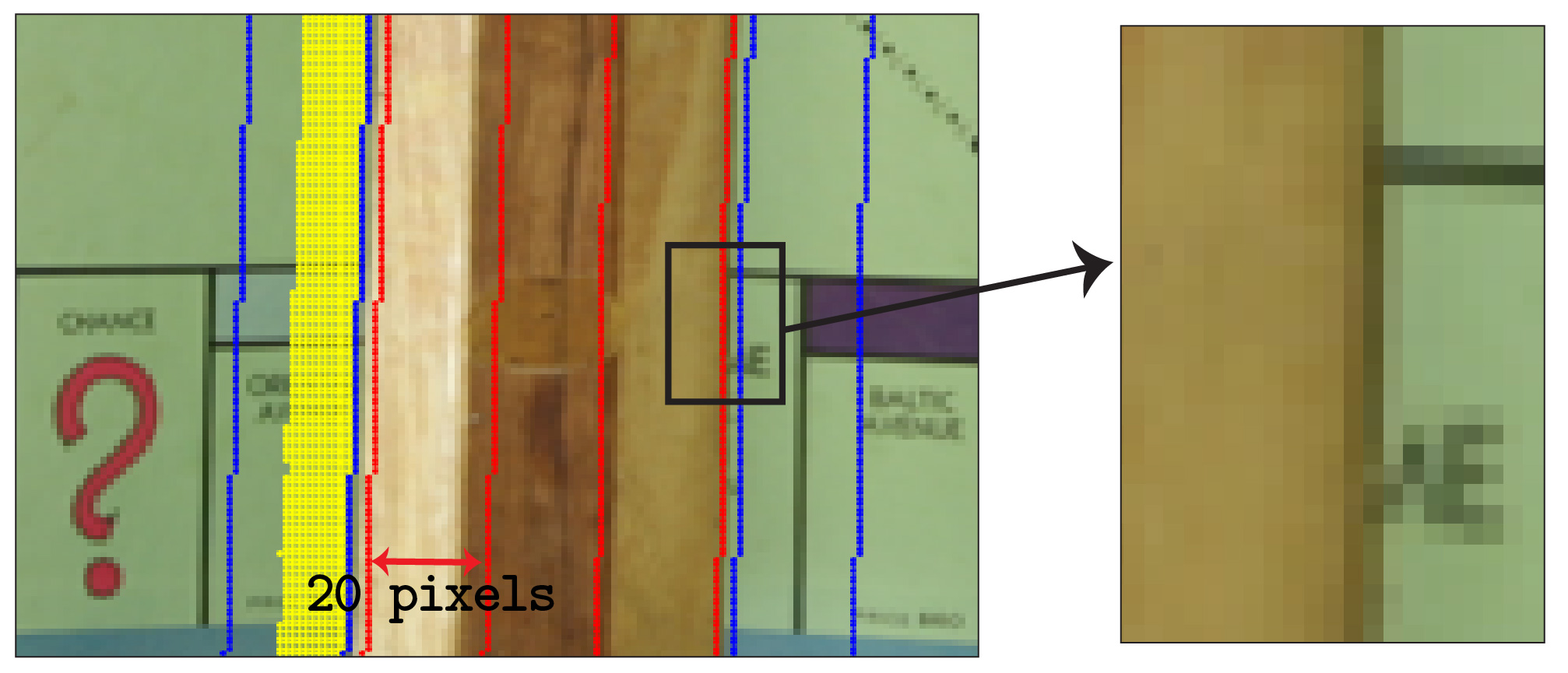}
    \caption{Evaluation area. We consider a region from 20 pixels left to 20 pixels right of the boundary except the center three pixels. See texts for discussion.}
    \label{fig:eval_area}
\end{figure}

We consider regions from 20 pixels left of the boundaries to 20 pixels right of the boundaries on the epipolar scanlines to be our region of interest. Due to lens blur and possible labeling errors, we exclude boundary $\pm1$ pixels from evaluation. Figure~\ref{fig:eval_area} shows an example. Regions between the red lines are foreground pixels we evaluate on, and regions between the blue lines are background pixels we evaluate on. Yellow dots are occluded pixels. The lens blur is demonstrated in the right column of Fig.~\ref{fig:eval_area}. Our occlusion boundary labels might be inaccurate up to one pixel, so we exclude the labeled boundary and the pixels immediate to the left and to the right of it.

\begin{figure}[h]
    \centering
    \includegraphics[width=0.9\textwidth]{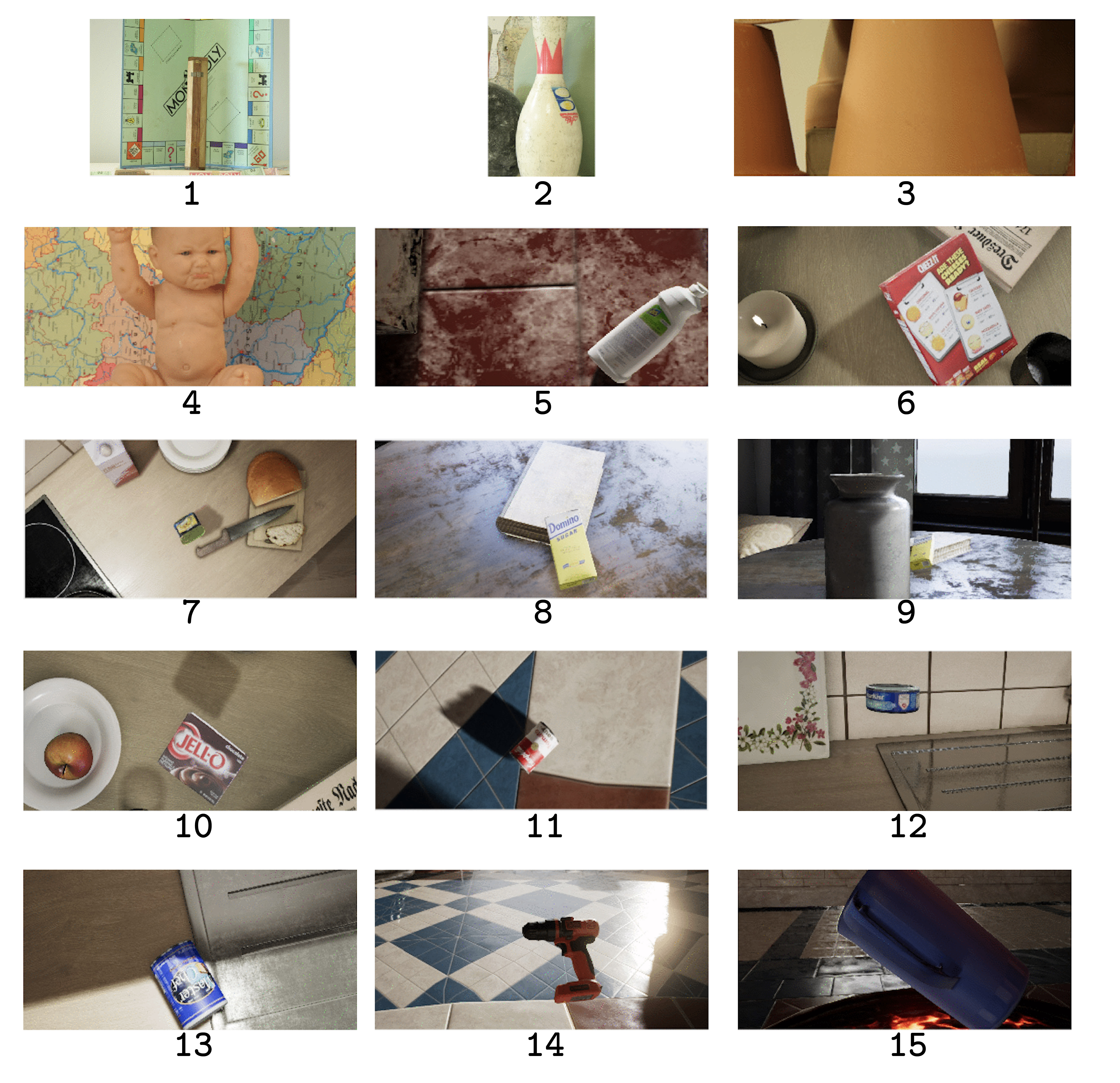}
    \caption{All fifteen test scenes we use to evaluate our method. Only the left stereo image is shown. Images 1-4 (from Middlebury) are of different resolutions. }
    \label{supfig:all_imgs}
\end{figure}

Figure \ref{supfig:all_imgs} shows all fifteen scenes in our dataset. The first four are cropped half-size Middlebury 2006 images~\cite{hirschmuller2007stereo, scharstein2007learning} and the rest eleven are from the Falling Things dataset~\cite{tremblay2018falling}. Table~\ref{tbl:result} shows the bad-4.0 result for each of the fifteen scenes (and the occlusion F1 scores are shown in the main paper).

\begin{table*}[t!]
\centering
\small
\setlength{\tabcolsep}{3pt}
\begin{tabular}{lccccccccccccccccc}
\toprule
Image & 1 & 2 & 3 & 4 & 5 & 6 & 7 & 8 & 9 & 10 & 11 & 12 & 13 & 14 & 15 & & Avg bad-4.0 \\
\midrule
SGM~\cite{hirschmuller2009matching} & 13.10 & 21.79 & 43.66 & 10.30 & 10.65 & 29.53 & 8.09 & 11.46 & 17.41 & 8.94 & 8.03 & 19.75 & 14.27 & 14.17 & 32.38 &  & 17.57 \\
BM-LR & 19.88 & 30.71 & 36.97 & 9.21 & 12.78 & 32.36 & 13.42 & 13.13 & 22.07 & 15.47 & 7.83 & 16.42 & 17.09 & 17.86 & 45.06 & & 20.68 \\
KZ~\cite{kolmogorov2001computing} & 31.26 & 13.15 & 49.97 & 3.84 & 3.53 & 18.86 & 2.10 & 3.22 & 12.52 & 1.62 & 2.57 & 9.99 & 24.60 & 57.40 & \textbf{29.05} & & 17.58 \\
HSM~\cite{yang2019hierarchical} & 1.69 & \textbf{3.35} & \textbf{0.25} & \textbf{1.80} & 0.99 & \textbf{2.92} & \textbf{0.00} & \textbf{0.00} & \textbf{2.59} & \textbf{0.89} & \textbf{0.00} & \textbf{5.69} & \textbf{2.38} & \textbf{3.18} & 63.08 & & \textbf{5.92} \\
Ours & \textbf{0.47} & 27.65 & 0.45 & 7.21 & \textbf{0.19} & 31.19 & \textbf{0.00} & 1.15 & 35.25 & 1.23 & 0.10 & 10.56 & 2.81 & 28.02 & 59.76 & & 13.74 \\
\bottomrule
\end{tabular}
\caption{Bad-4.0 result for each test image of Figure~\ref{supfig:all_imgs}.}
\label{tbl:result}
\vspace{-2mm}
\end{table*}

\begin{figure}[h!]
    \centering
    \includegraphics[width=0.88\textwidth]{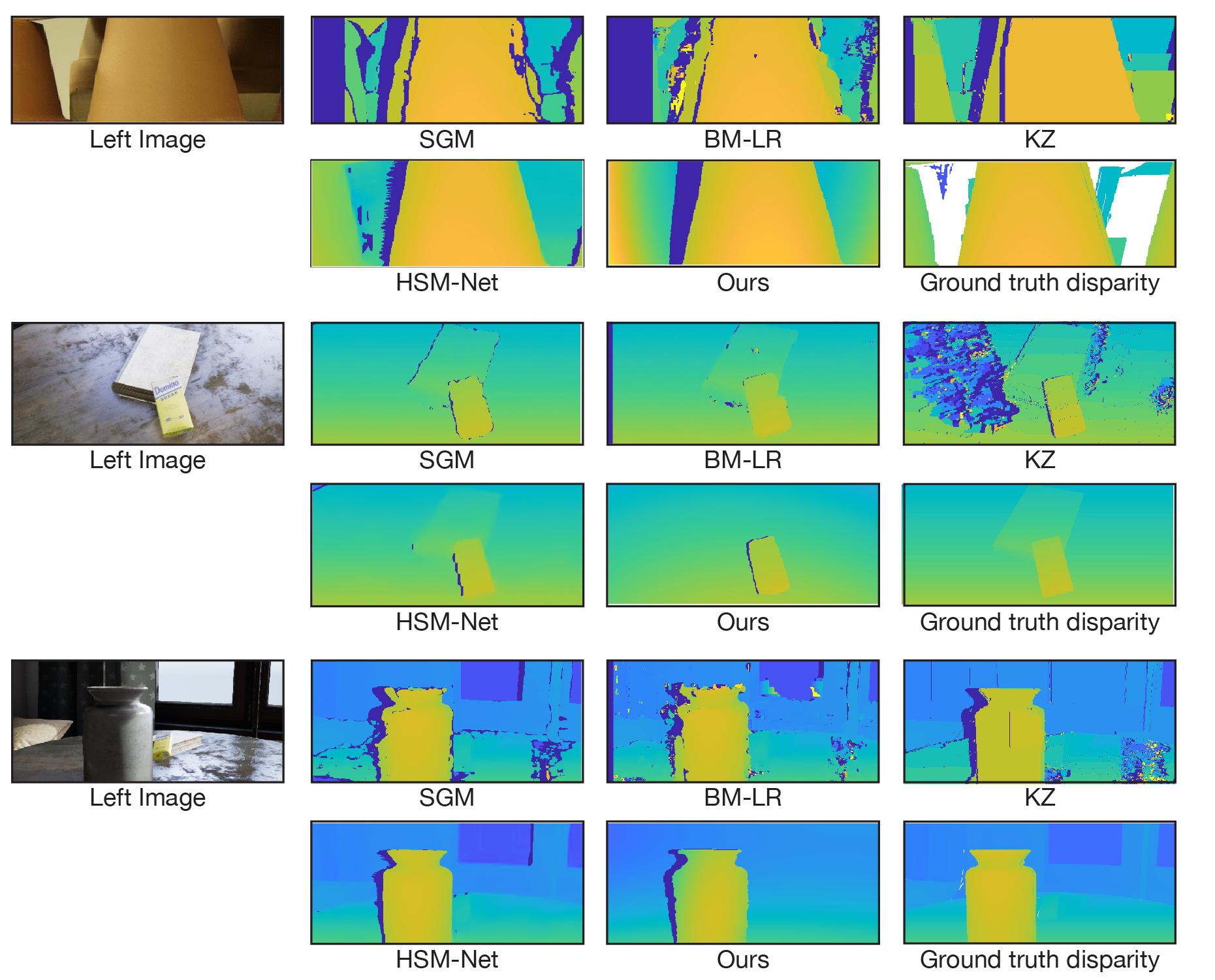}
    \caption{\textbf{Additional comparison.} The predicted disparity maps are shown for each method. Occluded regions are encoded in dark blue color. We use Matlab's implementation of SGM and BM-LR, which may not give predictions in the leftmost pixels that are visible only in the left image.}
    \label{supfig:more_comparison}
\end{figure}

Figure~\ref{supfig:more_comparison} shows three additional comparisons between our method and the baseline methods, each with a different occlusion handling approach. Dark blue regions are detected occlusions in each method and white regions are holes in the ground truth (The Middlebury stereo dataset has semi-dense ground truth data).

\begin{figure*}[h]
    \centering
    \includegraphics[width=0.95\textwidth]{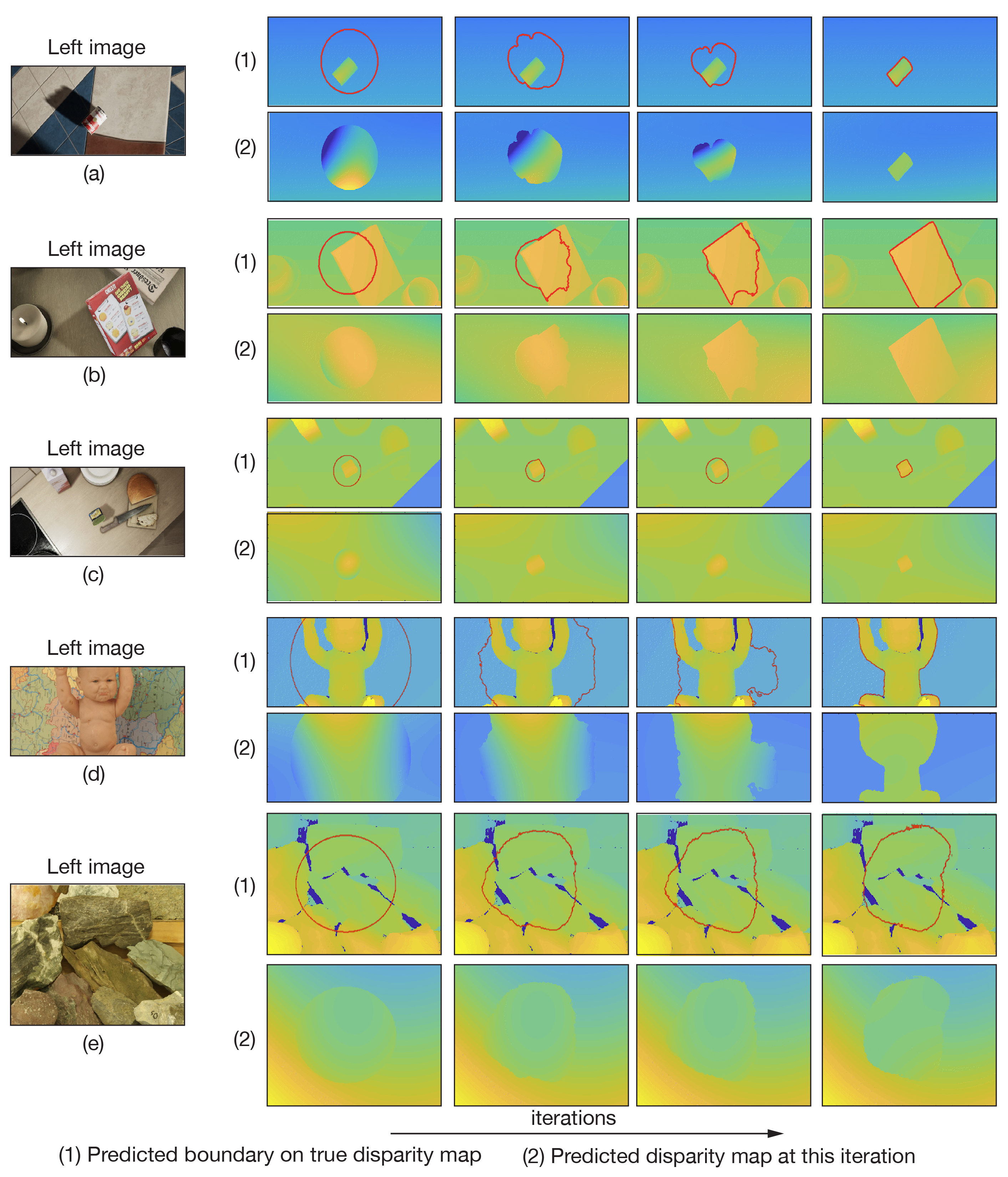}
    \caption{\textbf{Additional results.}  Three additional successful examples (a-c) and two failure cases (d-e) of Middlebury and Falling Things dataset~\cite{hirschmuller2007evaluation,scharstein2007learning,tremblay2018falling}. See text for discussion.}
    \label{supfig:more_results}
\end{figure*}

Figure~\ref{supfig:more_results} shows three additional successful results (a-c) and two failure cases (d-e). In (a-c), our model successfully segments the scene into two surfaces with precise boundaries. In (d), the pixels within the gaps between the doll's head and arms are mostly occluded. The matching cue is also weak there, especially with our simple matching cost. As a result, our model fails to group these regions with the rest of the background. 
(e) is another Middlebury 2006 scene that is not in our collection of test images, however, we use it here to show the model's behavior in a scene that differs substantially from a bi-layer one. The model tries to fit two quadratics to the scene to minimize the matching loss, and it ignores boundary and occlusion cues entirely. The predicted boundaries are not meaningful, but the disparity map still provides a coarsened approximation to the ground truth.

\clearpage
{
\bibliographystyle{IEEEbib}
\bibliography{refs}
}